\documentclass{article}
\usepackage[final]{neurips_2023}
\usepackage[utf8]{inputenc}
\usepackage[T1]{fontenc}
\usepackage{hyperref}   
\usepackage{url}        
\usepackage{booktabs}   
\usepackage{amsfonts}   
\usepackage{nicefrac}   
\usepackage{microtype}  
\usepackage{xcolor}     
\usepackage{graphicx}
\usepackage{tcolorbox}
\usepackage{amsmath}
\usepackage[utf8]{inputenc}
\usepackage{siunitx}
\usepackage{float}
\usepackage{multirow}

\newcommand\norm[1]{\lVert#1\rVert}
\newcommand{\etal}{\textit{et al}. }

\definecolor{average}{HTML}{006101}
\definecolor{ideal}{HTML}{286ba8}
\definecolor{sample}{HTML}{ff3f37}
\usepackage{wrapfig}

\definecolor{ps}{RGB}{128,0,128}

\title{Don't Miss Out on Novelty:
 Importance \\of Novel Features for Deep Anomaly Detection}

\author{%
Sarath Sivaprasad,\, \hspace{0.2cm} Mario Fritz \\
CISPA Helmholtz Center for Information Security \\
\small{\texttt{\{sarath.sivaprasad, fritz\}@cispa.de}} \\
}

\begin{document}

\maketitle

\begin{abstract}

Anomaly Detection (AD) is a critical task that involves identifying observations that do not conform to a learned model of normality.
Prior work in deep AD is predominantly based on a \textit{familiarity hypothesis}, where \textit{familiar} features serve as the reference in a pre-trained {\it embedding space}.
While this strategy has proven highly successful, it turns out that it causes consistent false negatives when anomalies consist of truly \textit{novel} features that are not well captured by the pre-trained encoding. We propose a novel approach to AD using explainability to capture such novel features as unexplained observations in the {\it input space}. We achieve strong performance across a wide range of anomaly benchmarks by combining familiarity and novelty in a hybrid approach.
Our approach establishes a new state-of-the-art across multiple benchmarks, handling diverse anomaly types while eliminating the need for expensive background models and dense matching.
In particular, we show that by taking account of novel features, we reduce false negative anomalies by up to 40\% on challenging benchmarks compared to the state-of-the-art. Our method gives visually inspectable explanations for pixel-level anomalies. 

\end{abstract}

\section{Introduction}

Anomaly detection (AD) is a crucial task that involves identifying test samples that deviate significantly from the normal training samples. 
In real-world applications, the occurrence of anomalies can lead to severe consequences, to the extent that AD has been identified as a critical component in improving organizational security under catastrophic AI risks~\cite {hendrycks2023overview}.
Anomaly detection methods are used across diverse domains, like quality control in manufacturing~\cite{zipfel2023anomaly}, medical imaging for early disease diagnosis~\cite{fernando2021deep}, enhancing security and surveillance systems~\cite{sodemann2012review}.
Across the different applications, image anomalies are broadly classified into two: semantic anomaly, a sample outside the `normal' semantic distribution, and sensory anomaly, caused by unexpected pixel-level aberrations in an otherwise normal sample. In deep anomaly detection, these anomaly types are often handled with specialized approaches~\cite{jiang2022survey}.

\begin{figure}
  \centering
\includegraphics[width=0.8\linewidth]{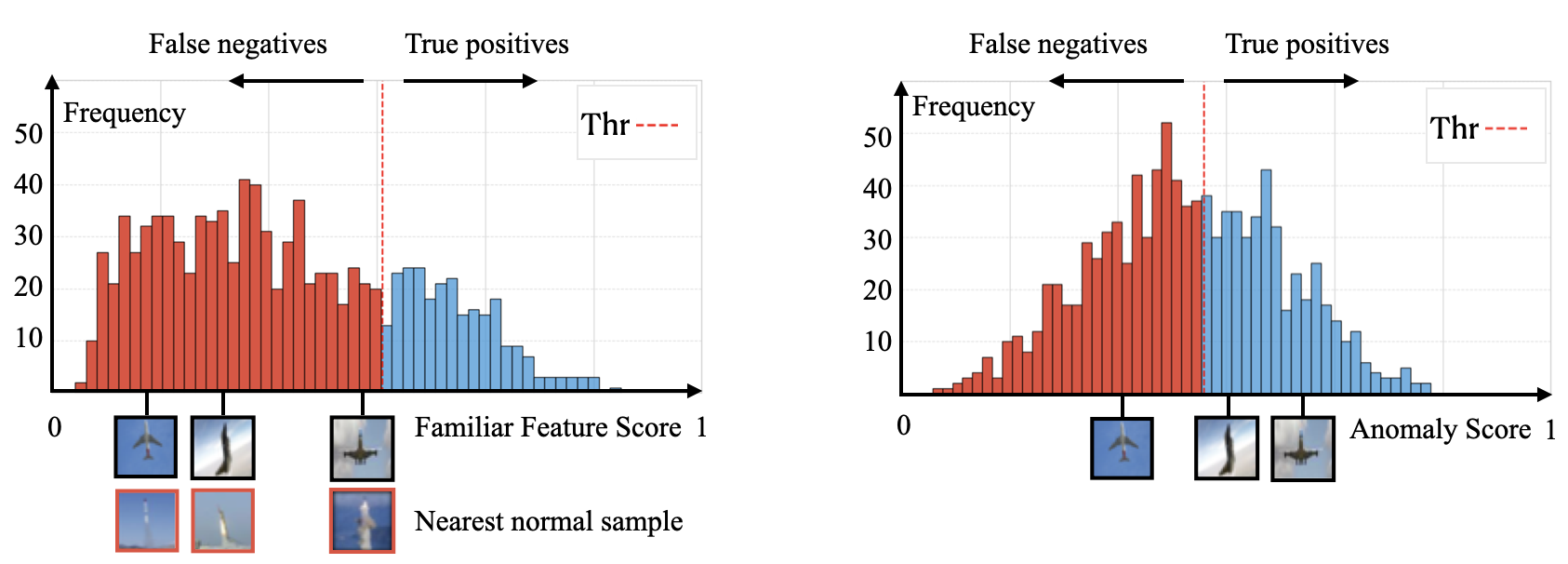}
  \caption{Example anomaly detection problem on CIFAR-100: rocket$=$normal; CIFAR-10: all classes$=$anomaly. Familiarity only causes false positives as rockets are close to planes in the embedding space and hence have a too-low familiarity-based anomaly score. (Right) our method corrects more false positives by taking novel features into account.}
  \label{fig:teaser}
  \vspace{-0.8mm}
\end{figure}
  \vspace{-1.8mm}

Prior work in deep AD for images is predominantly based on the familiarity hypothesis, where the anomaly is identified by the lack of \textit{familiar} features in them~\cite{familiarity}.
Familiar features are the set of features the neural encoder has learned to represent meaningfully in the representation space. 
Inventive methods have been proposed to learn feature spaces where anomalies can be characterized by a lack of familiar features. The state-of-the-art AD method uses feature representation of a pretrained ViT backbone fine-tuned to classify normal samples from samples generated using a diffusion model prematurely early stopped while approximating the normal distribution~\cite{mirzaei2023fake}.

Relying solely on the set of familiar features leads to two major issues in deep AD.
Firstly, neural networks show 'unreasonable' generalize well beyond the training data~\cite{zhang2021understanding}, often showing invariance in representation to even OOD samples~\cite{jacobsen2018excessive}.
This excessive invariance of representation well beyond the train distribution can lead to false negatives in familiarity based AD.
Secondly, this paradigm does not account for anomalies caused by truly novel features not being represented meaningfully in the feature space through the learned encoding, also leading to false negatives~\ref{fig:teaser}.

While significant work has gone into countering the excessive generalization~\cite{mirzaei2023fake, cohen2022transformaly, tack2020csi} the latter issue needs a different modelling and strategy.
In particular, current successful pre-trained embeddings might capture some features poorly or not at all.
Moreover, solving the excessive generalization often involves making assumptions on the nature of anomalies~\cite{hendrycks2018deep} or generating complex distribution~\cite{mirzaei2023fake} as outlier samples to control generalization.

We propose a novel approach for AD that addresses both these key issues by jointly modelling the lack of familiarity and presence of novelty towards anomaly detection.
We use the features extracted by the encoder to compute familiarity and capture novel features as unexplained observations in the input space.
A faithful explanation of the encoding enables inspection of features that were not meaningfully interpreted by the encoder. In this work, we use $\operatorname{B-cos}$ networks ~\cite{Boehle2022CVPR} to summarize the encoder into a faithful explanation of the encoding.
We show that accounting for novel features for AD reduces the reliance on a complex outlier distribution to control generalization. 
While the two scores are not mutually exclusive, our experiments show that the latter adds to AD performance~\ref{fig:teaser_1}. 

We evaluate the method across multiple benchmarks and establish new state-of-the-art in most of the evaluated benchmarks. 
In particular, we show that by taking account of novel features, we reduce false negative anomalies by up to 40\% across challenging benchmarks.
Our experiments show that joint modelling eliminates the need for expensive outlier samples and dense matching to improve AD performance.
For sensory anomaly, the explanation is traced to the input, giving visually inspectable explanations. The method adapts to different anomaly types by varying the layer at which the novelty is computed.
Since early layers of the backbone pre-trained on large datasets are frozen while training for AD tasks, we compute novelty with respect to these features for detecting high-level semantic novelty.
In short, we make the following contributions:

\begin{itemize}
    \item We introduce the concept of familiar and novel features in a test input in the context of anomaly detection. We propose using a lack of faithful explanations to capture the novel features.
    
    \item We propose a joint model for AD that accounts for the lack of {\it familiarity} and the presence of {\it novelty} in an input sample. We evaluate this model on sensory level and semantic level anomaly detection by detecting novelty at different hierarchies of the neural network. It gives visually inspectable explanations for sensory anomalies.
    
    \item We show that the proposed method reduces the false negatives by accounting for novel features. Further experiments show that the proposed method also reduces reliance on the training outliers. 
\end{itemize}

\section{Related Work}
\label{Related Work}

This section discusses the methods developed to improve deep AD in the prior art. Most deep AD methods first train an encoder to learn a representation for normal samples and then use this representation to compute the anomaly score for a test sample~\cite{hojjati2022self, han2022adbench}. Hence, we categorize the popular methods as improving feature representation and test time detection. The survey shows that anomaly detection (though often synonymous with novelty detection) has often been solved by detecting the absence of familiarity.

\begin{wrapfigure}{r}{0.5\textwidth}
  \centering
  \includegraphics[width=0.48\textwidth]{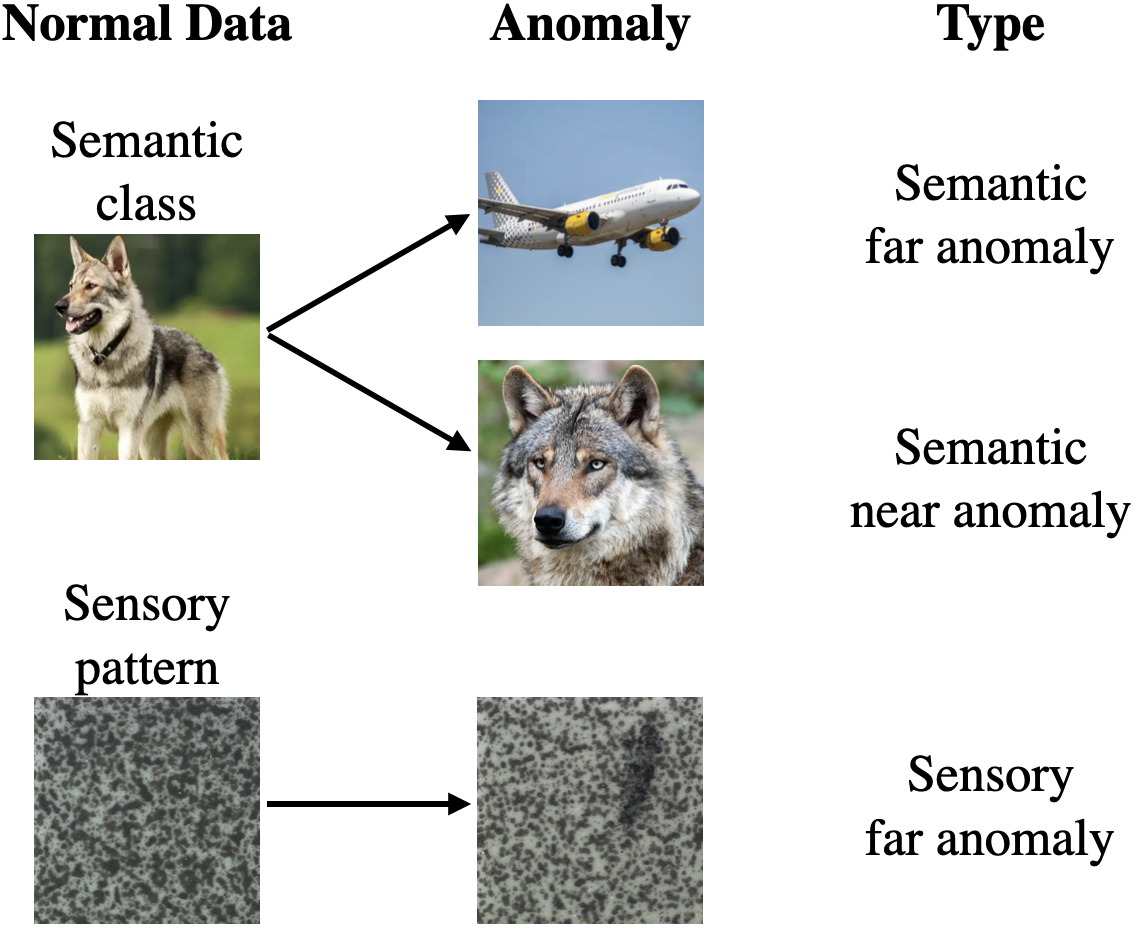}
  \caption{Illustration of predominant anomaly types considered in prior work.}
  \label{fig:anomaly types}
\end{wrapfigure}

\textbf{Improving feature learning for AD: }
Deep AD is generally achieved in a two-step approach ~\cite{mirzaei2023fake, bergmann2020uninformed, reiss2021panda, salehi2021multiresolution}. The first step learns a feature representation. Perera~\etal notes that using NN's in AD aims to learn robust feature space to define normalcy ~\cite{perera2019learning}. Bergman~\etal shows that using backbone pre-trained on large datasets substantially improves AD performance~\cite {bergman2020deep}. Fort~\etal demonstrates that transformers (pre-trained on large datasets) can significantly improve OOD tasks across different data modalities~\cite{fort2021exploring}. 
Investigating the trend of finding better representation space for AD Reiss~\etal provides theoretical and empirical evidence to show that AD cannot improve indefinitely by increasing the expressiveness of networks~\cite{reiss2023no}. In fact, they show that there is a trade-off between expressiveness of features and sensitivity to anomalies.

AD performance has been shown to improve by controlling the generalization of the NN encoder using fine-tuning with, controlled outlier exposure, using real outliers, random images from the internet, or other samples from other datasets \cite{hendrycks2019using,fort2021exploring}. Using GANs to generate outliers\cite{kong2021opengan,pourreza2021g2d} shows further improvement over real images. Mirzaei \etal  \cite{mirzaei2023fake} use a prematurely stopped SDE model to generate background samples at the boundary of the distribution. A desirable property of AD is a reduced reliance on background class and minimal assumptions on the nature of anomaly. Our method shows robust performance with simple background approximation.

\textbf{Test time detection methods: }
Popular detection methods for OOD detection and AD are: maximum softmax probabilities~\cite{hendrycks2016baseline}, Local Outlier Factor~\cite{lin2019deep},  Gaussian Discriminant Analysis~\cite{xu2020deep} or nearest neighbor ~\cite{bergman2020deep} to compute the similarity between the representations of normal samples and the test sample. Reiss \etal observe that these methods are opaque and non-interpretable. As anomalies are ambiguous, it is necessary to give explicit reasoning behind the criteria for detection. Our method explains the anomaly score when the anomaly is spatially local.

\textbf{Anomaly detection methods for sensory anomalies: } According to the different distribution shifts that cause them, anomalies 
are divided into sensory and sensory anomalies and semantic AD (Figure ~\ref{fig:anomaly types}) \cite{NEURIPS2022_d201587e}\cite{jiang2022survey}.
As sensory anomalies contain dense familiar features, its challenging to tackle via familiarity hypothesis~\cite{mvtech}. Most methods use locally sensitive dense feature extractors such that a novelty in input can come only at the cost of lost familiarity~\cite{roth2021total, cohen2020sub}. With the increasing number of normal samples, the memory bank becomes exceedingly large, with it both the inference time and memory required. Roth \etal uses a coreset sub-sampling to reduce this effect~\cite{roth2021total}. Jiang~\etal surveys methods for visual sensory anomalies and notes that most methods tuned for sensory anomalies perform poorly in detecting semantic outliers\cite{jiang2022survey}. Accounting for novelty in input could be key in bridging this gap in performance. 

\textbf{The familiarity hypothesis: } For a regular NN, activation in the last layer for a novel class sample is usually much smaller than samples from training data. Vaze~\etal suggests that this difference can be a good open-set-recognition~\cite{vaze2022openset}. Neural encoders give dense representations when familiar features are present in the input and fail to give an equally dense representation for samples with novel features~\cite{tack2020csi}. Dietterich~\etal formulates this as the familiarity hypothesis and argues that familiarity-based detection is an inevitable consequence of representation learning in AD~\cite{familiarity}. 
Previous efforts to create a hybrid model for AD unify generative modelling approaches for regular training data and discriminating with respect to negative training data~\cite{grcic2022densehybrid}. Building on the familiarity hypothesis, we propose augmenting the feature familiarity score with a score that accounts for novelty in the input for AD.

\begin{figure}[t]
  \centering
\includegraphics[width=\linewidth]{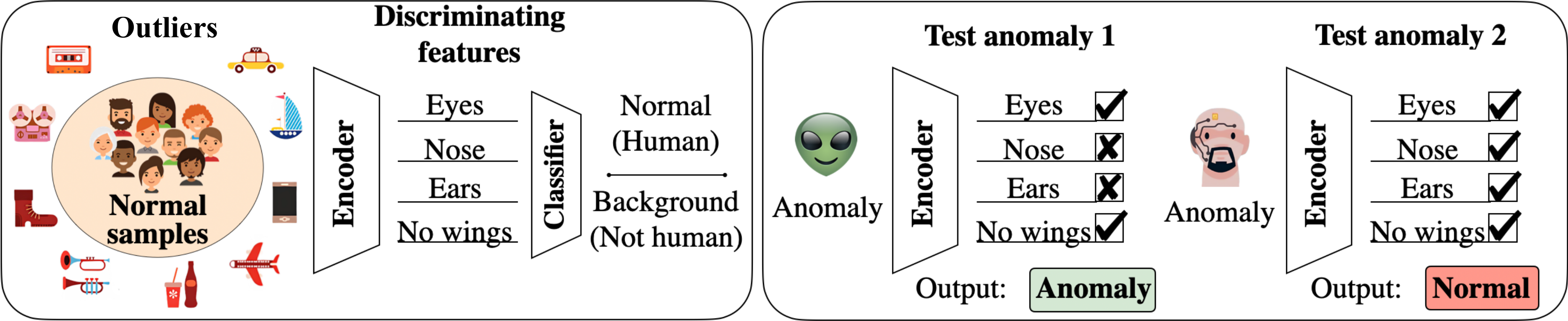}
  \caption{Feature learning based AD methods succeed by detecting the presence and absence of \textit{familiar} features in the test sample. Familiar features are the features the encoder learns to discriminate the normal samples from the outliers. The detection method fails for samples with \textit{novel} features that the encoder is not trained to represent in the feature space.}
  \label{fig:teaser_1}
\end{figure}

\section{Joint Model for familiarity and novelty based Anomaly Detection}

\begin{figure*}[t]
  \centering
    \includegraphics[width=0.9\linewidth]{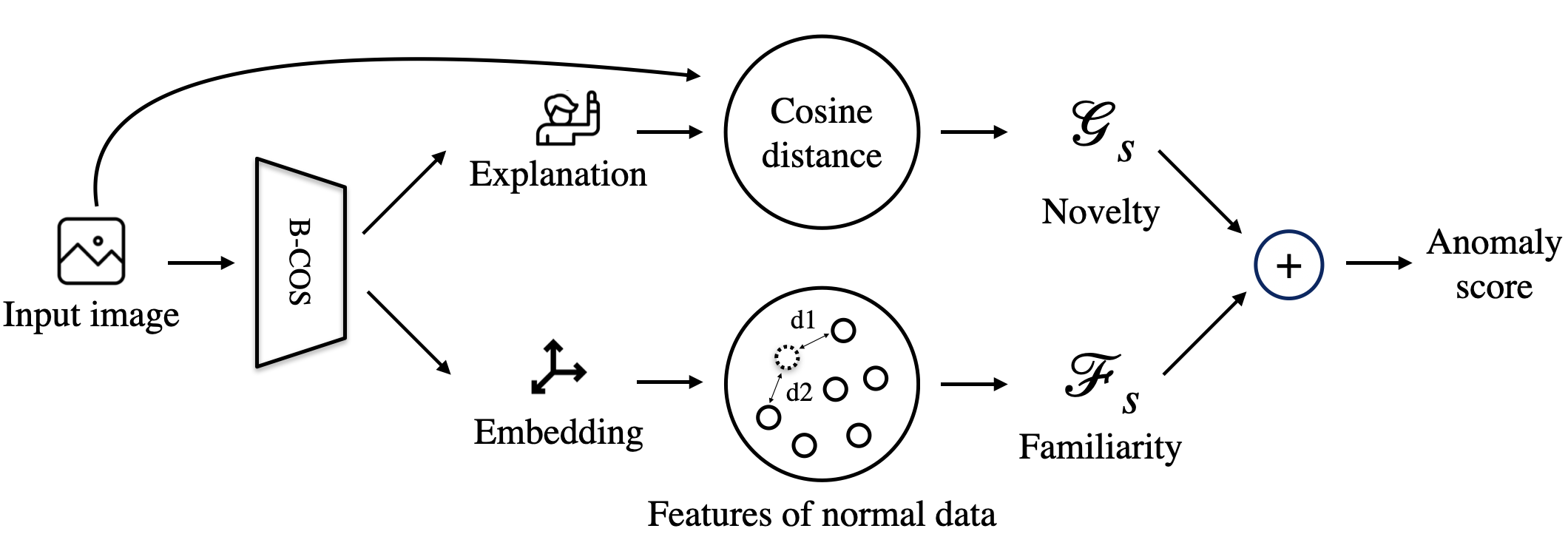}
  \caption{Figure shows the proposed pipeline. The top portion computes Explanation-based Novelty Score (ENS) and the bottom branch computes Familiar Feature based anomaly Score (FFS). The final score, novelty accounted anomaly score is a combination of both.}
  \label{fig:model}
\end{figure*}

In this section, we describe the proposed method to jointly model the lack of familiarity and the presence of novelty for anomaly detection. We propose computing a joint anomaly score ($\mathcal{A}$) for a test sample $x_\mathrm{test}$ as the weighted sum of the familiarity and novelty scores of the sample.

\begin{equation}
\mathcal{A}(x_\mathrm{test}) = \mathcal{F}_s(x_\mathrm{test}) + w.\mathcal{G}_s(x_\mathrm{test})
\label{eq:fam}
\end{equation}

where $\mathcal{F}_s$ is the familiarity score and $\mathcal{G}_s$ the novelty score associated to the test sample $x_\mathrm{test}$. $w$ is the weight to balance the scale. The pipeline of the proposed AD method is shown in Figure~\ref{fig:model}. The figure shows how, given a test sample, the familiarity and novelty scores are computed using the train features and the encoder explanation, respectively.

The detection is similar to prior art with a two-step approach. An encoder $F$ is trained on a related task (a task close to AD): discriminating normal train samples from the outliers. We define familiar features for an AD task as the set of all features the fine-tuned encoder ($F(\theta)$) has learnt to encode meaningfully in the feature space. AD score of a test sample in the familiarity branch is solely computed using the familiar features in the test sample and familiar features in normal train samples. However, the test sample can contain features outside this set that make it an anomaly. We define novel features as the features in the test sample that are not in the familiar feature set. In other words, they are the test input features that the encoder does not represent meaningfully in the feature space.

\label{FFS}
\subsection {Familiar Feature Based Anomaly Score (FFS)}
In order to detect an anomaly by the lack of familiar features in a test sample, we use a mechanism similar to prior art. We first train an encoder $F$ to discriminate a subset of normal data $n$ from a subset of 
outliers $b$ to get parameter $\theta$. Using $F(\theta)$ as encoder, we compute the distance between features of the test sample ($F(x_\mathrm{test})$) and the train normal samples ($F(\theta,n_i)\forall n_i\in n $). We call this measure a Familiar Feature based anomaly Score (FFS). 

We compute $F(\theta ,n_i) \forall n_i \in n$ and store them as the rows of matrix $M$. The FFS score $\mathcal{F}_s$ for input $x_\mathrm{test}$ is computed as the sum of distances of the test feature to the two nearest train features following ~\cite{mirzaei2023fake}, details of the choice of hyperparameters is in Appendix 3. FFS score increases with the lack of familiar features in the test sample.

\begin{equation}
\mathcal{F}_s(x_\mathrm{test}) = \norm{F(\theta,x_\mathrm{test}) - M_{0}}  +  \norm{F(\theta,x_\mathrm{test}) - M_{1}} 
\label{eq:fam}
\end{equation}

where $M_{0}$ and $M_{1}$ corresponds to the rows in $M$ closest to the test feature vector. This method requires the computation and storing of all train normal sample representations. Prior art like \cite{mirzaei2023fake} reports FFS as the anomaly score.

\subsection{Explanation Based Novelty Score (ENS)}
\label{ENS}

Novel features are the features in the test input $x_\mathrm{test}$ that are not meaningfully represented by the encoder $F(\theta)$. The score is not dependent on the train normal features (elements of $M$). To capture the novel feature in a test input, we use the explanation of a $\operatorname{B-cos}$ network. An encoder built with $\operatorname{B-cos}$ operator generates a reliable explanation of its computation. $\operatorname{B-cos}$ networks are neural networks where the linear layers are replaced by $\operatorname{B-cos}$ layers. For more details on the formulation and training of the networks, we refer the reader to ~\cite{Boehle2022CVPR}. Operation of a $\operatorname{B-cos}$ layer at a node for an input $x$ and parameters $w$ leading to the node is given by

\begin{equation}
\operatorname{B-cos}(x;w) = \norm{x} \cdot \norm{w} \cdot \cos(\angle(x,w))^{B} \cdot  \mathrm{sign}(\cos(\angle(x,w)) )
\label{Bcos-equ}
\end{equation}

Where $B$ is a hyper-parameter that influences the extent to which alignment between $x$ and $w$ contributes to the magnitude of the output. Using the $\operatorname{B-cos}$ transforms instead of linear transform removes the need for other explicit non-linearity while training the network. Hence, the only non-linearity in the network is dependent on the input. Given an input, $\operatorname{B-cos}$ network collapses into a single linear transform that faithfully summarises the entire model computations. Moreover, the $\operatorname{B-cos}$ transform introduces alignment pressure on the weights during optimization. For the output of a node to be high, it requires that the input is aligned to the incoming parameters of the node ($\cos(\angle(x,w))$ is high). When the output of the network is high, the summarized linear layer is highly aligned to the input. Hence, using a $\operatorname{B-cos}$ network helps generate a faithful explanation of the decision aligned to the parts of the image that contribute to activating the network's output.

We use the lack of this explanation as evidence of the presence of novel features. We quantify the size of the novel feature set using the cosine distance between $x_\mathrm{test}$ and the explanation of the encoder explanation and use this to compute novelty score. $\operatorname{B-cos}$ encoder explanation gives a reliable summarization of network computation. 

We approximate novel features as that portion of the input features that the pre-trained $\operatorname{B-cos}$ encoder cannot align to. It corresponds to the feature in the input that the encoder fails to explain in the context of the decision. For a given input $x$ Böhle \etal denotes the explanation of an $L$ layer neural network as $\theta_{1\rightarrow L}(x)$. We compute the Explanation based Novelty Score (ENS) denoted by $\mathcal{G}_s$ for a test input $x_{\mathrm{test}}$, for an encoder with parameters $\theta$ as

\begin{equation}
\mathcal{G}_s(x_{\mathrm{test}}) = 1 - \cos(\angle( \theta_{1\rightarrow L}(x_{\mathrm{test}}), x_{\mathrm{test}}))
\label{Novelty}
\end{equation}

Note that the $\mathcal{F}_s$ score is computed using features alone while $\mathcal{G}_s$ does not rely on the encoded normal features. Finally, we compute the joint anomaly score as the sum of normalized familiarity and novelty score.

\subsection{ Adapting to Anomaly Types}
\label{feature_exp}

Since we use a backbone pre-trained on large data, the initial layers derive a wide range of features. With frozen initial layers, it becomes meaningful to check for novel features higher up in the neural layer hierarchy with respect to these features. For a test sample, Böhle~\etal uses $\theta_{1\rightarrow L}(x_{\mathrm{test}})$ to visualize the explanation of the decision. The layers are collapsed from the input to the output node of a classifier of $L$ layers. We modify this formulation to capture the the portion of features that are explained given the final encoding, instead of computing the portion of input that explains the decision. That is, we compute $W_{l\rightarrow L}$ where $l$ is the layer at which we evaluate the novelty and $L$ is the final layer. Novelty of feature $F_{i}$, output by layer $i$ is computed as,

\begin{equation}
\mathcal{G}_s(f_i) = 1 - \cos(\angle( \theta_{(i+1)\rightarrow L}(f_i), f_i))
\label{our_method_eq}
\end{equation}

For sensory anomalies, the value of $i$ is one, and  Equation~\ref{our_method_eq} becomes similar to the formulation in the prior art. Here, the choice of layer is a hyper-parameter to adapt to anomalies at different semantic levels.

\section{Experiments and Results}

This section discusses the four experiments to evaluate the effect of accounting for novel features for computing anomaly scores. In the first experiment, we demonstrate the efficacy of the proposed framework on different AD benchmarks. We benchmark our method on eight different datasets to evaluate its overall effectiveness. The benchmark shows the efficacy of the method across different anomaly types, the sensory anomaly, and visually near and far semantic anomaly. We then present an analysis to show the effectiveness of the joint model in reducing false negatives. The second experiment evaluates the performance of the proposed novelty capture method in case of sensory anomalies and shows how this can give visually inspectable explanations. In the third experiment, we show an ablation to show the effect of FFS and ENS scores towards anomaly prediction. In the final experiment, we showcase how our method helps reduce reliance on outlier classes.

\subsection{Benchmarking across different AD tasks}
For benchmarking on different types of anomalies, we choose eight different datasets across the three anomaly types. The effective number of datasets evaluated is much more, as an evaluation is done for each class in each of the datasets as the normal class. Furthermore, for each dataset, for every normal class, we fine-tune a $\operatorname{B-cos}$ ViT backbone pre-trained on the ImageNet-1K dataset with a two-class classification head. The two class classification head correspond to normal and outlier classes. We use data samples from the normal approximation of the normal class as the anomaly. The proposed method gives a hyper-parameter to control the semantic level at which the anomaly is computed (variable $i$ in Section ~\ref{feature_exp}). For a fair comparison, we do not tune this for each dataset to optimize the performance. Across benchmarks, we use $i = 0$ for the sensory anomaly (the anomaly is in pixel level), $i = 6$ for all far anomaly (the anomaly is at low-level visual features), and $i = L-1$ for semantic near anomaly (the anomaly is at high-level semantics). The further exploration of this parameter is left for future work.

\begin{table*}[h!]
    \centering
    \resizebox{\textwidth}{!}{%

    \begin{tabular}{l *{4}{l}|l l l | l | l}
    \toprule
    \multirow{2}{*}{Method} & \multicolumn{8}{c}{Datasets} \\
    \cmidrule{2-10}
    \multirow{2}{*}{} & \multicolumn{4}{c}{Semantic near AD} & \multicolumn{3}{c}{Semantic far AD} & \multicolumn{1}{c}{Sensory AD}\\

    \cmidrule{2-10}
    & CIFAR-10 & CIFAR-100 & Flowers & Birds & FGVC  & Cars & C10-100 & MVTec & Average \\
    \midrule
    CSI \cite{tack2020csi} & 94.3 & 89.6 & 60.8 & 52.4 & 64.6 & 66.5 & 76.1 & 63.6 & 71.0 \\
    MSAD(ViT) \cite{reiss2023mean} & 94.1 & 93.0 & 98.6  & 93.3 &  81.3 & 85.7 & 79.5 & 85.5 & 88.8 \\
    Transformaly \cite{cohen2022transformaly} & 98.3 & 97.3 & 99.9 & 97.8 & 84.0  & 86.7 & 82.5 & 87.9 & 91.8 \\
    PANDA \cite{reiss2021panda} & 96.2 & 94.1 & \underline{94.1}  & 95.3 & 77.7 & 87.6 & 76.8 & 86.5  & 88.5 \\
    PatchCore \cite{roth2021total} & 67.2 & 64.1 & 74.8 &  58.1 & 67.8  & 78.3  & 67.2 & \textbf{99.1} & 72.1 \\
    FITYMI \cite{mirzaei2023fake} & \underline{99.1} & \underline{98.1} & \textbf{99.9} & \underline{98.5}  & \underline{88.7} &  \textbf{90.8} & \underline{89.4} & 86.4  & \underline{93.8} \\
    \midrule
    Our method & \textbf{99.3} & \textbf{98.5} & \textbf{99.9} & \textbf{98.7} & \textbf{89.3}  & \underline{90.5} & \textbf{91.1}  & \underline{89.3} & \textbf{95.3} \\
    \bottomrule
    \end{tabular}}
    \caption{The performance of the proposed method for semantic anomaly detection methods (AUROC) in the AD setting on different datasets. The best performance of the best-performing model is bold, and the second-best method is underlined.}
    \vspace{-0.5cm}
    \label{benchmark}
\end{table*}

Table \ref{benchmark} shows the performance of various prior-art benchmarked against the proposed method. The performance of some of the datasets like CIFAR-10 and Flowers are saturated (above 99\%). On average, our model outperforms the next best model by more than 2.5\%.

On the more challenging benchmark of near semantic AD (CIFAR-10 vs CIFAR-100~\cite{mirzaei2023fake}, where the closest classes of CIFAR-100 corresponding to each class of CIFAR-10 is picked for test), our method outperforms the familiarity based method by 1.0\% establishing the new SOTA ( details of experiment setup in Appendix 2). Furthermore, the table shows an interesting trade-off of performance across the two challenging tasks of near semantic AD and sensory anomaly the best-performing methods are different. While PatchCore outperforms FITYMI by a margin of more than 13\% on MVTec, FITYMI outperforms PatchCore by more than 22\% on near-semantic AD. This shows how one method is tuned for semantic anomalies and the other for sensory anomalies. Using novelty on the familiarity gives a more consistent performance across the two tasks. The gap in performance across the two tasks shows a scope for improvement in computing novelty.

\begin{wrapfigure}{r}{0.5\textwidth}
  \centering
  \includegraphics[width=0.48\textwidth]{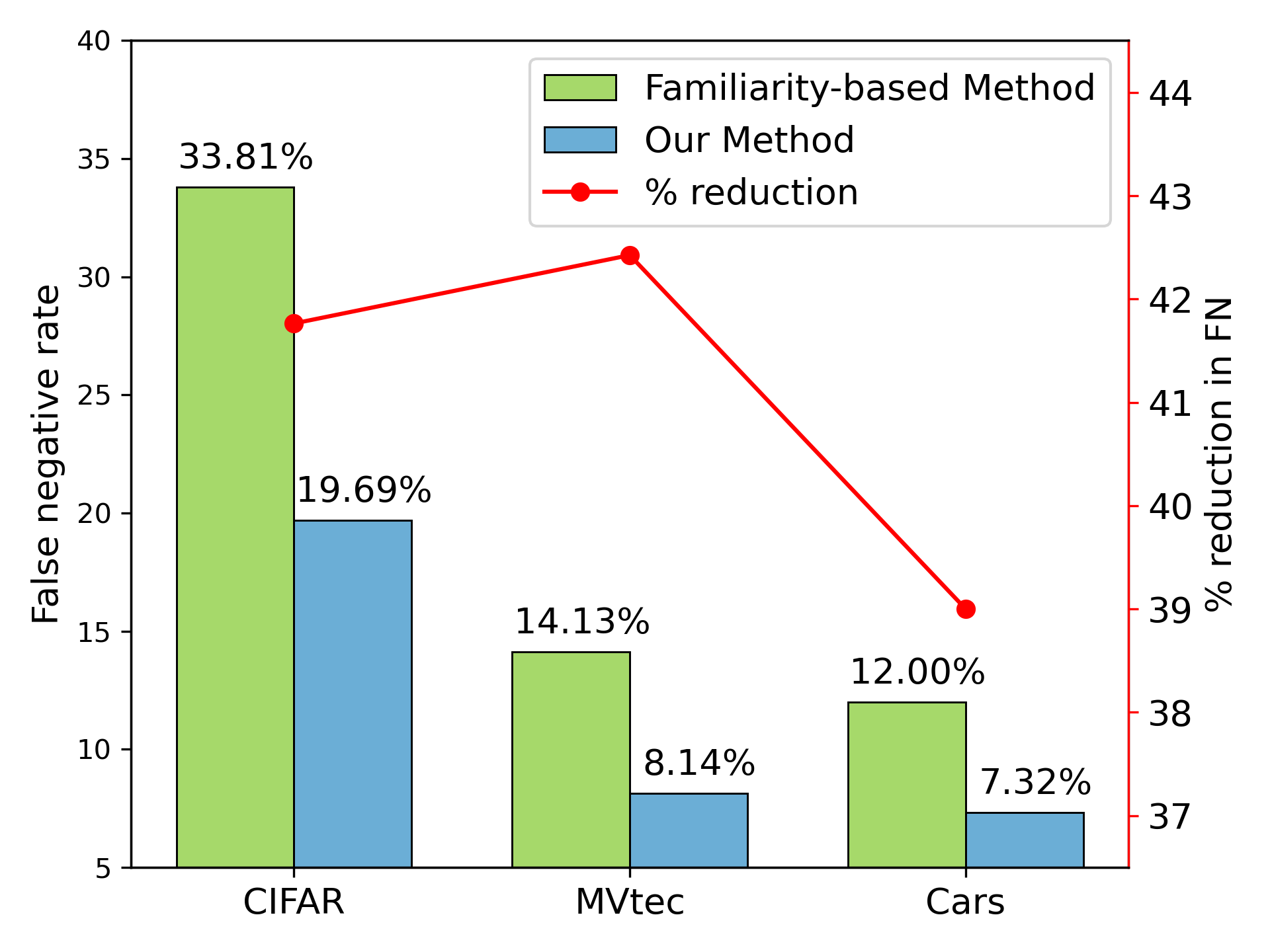}
    \caption{Comparing the false positives across different datasets with and without accounting for novelty. Y-axis: left shows the rate of FP, and the right side shows the \% reduction in FP.}
    \label{fig:FPR}
\end{wrapfigure}

\paragraph{Reduced False negatives:} 
Methods that use only familiar features will find it challenging to detect anomalies caused by truly novel features, producing false negative predictions. In this experiment, we ablate the effect of familiarity and novelty branches to evaluate the role of incorporating novelty into the scoring mechanism in the false positive rate on three benchmarks: Stanford-Cars (semantic far AD), Cifar10-100 (semantic near AD), and MVTec (sensory AD). Note that FFS scoring is similar to ~\cite{mirzaei2023fake} and compared with the addition of ENS to compute novelty. Furthermore, the False Negative rate is an important characterization of an AD method in high-risk applications. To compute the same, we convert the anomaly score into classification. We use an oracle to find the optimum threshold for each class of each dataset. The results show that (Figure~\ref{fig:FPR}) accounting for novelty in input reduces the number of false positives by about 40\% across the different anomaly types. This validates the hypothesis in \cite{familiarity} that AD by relying solely on familiar features, can lead to missing the anomalies caused by novel features. The teaser figure also shows the effect of adding the novelty score to the FFS score on a configuration of the C10-100 dataset, where a class (rocket) is the normal class, and all test samples of C10 are anomalies.

\begin{figure*}[t]
  \centering
  \includegraphics[width=\linewidth]{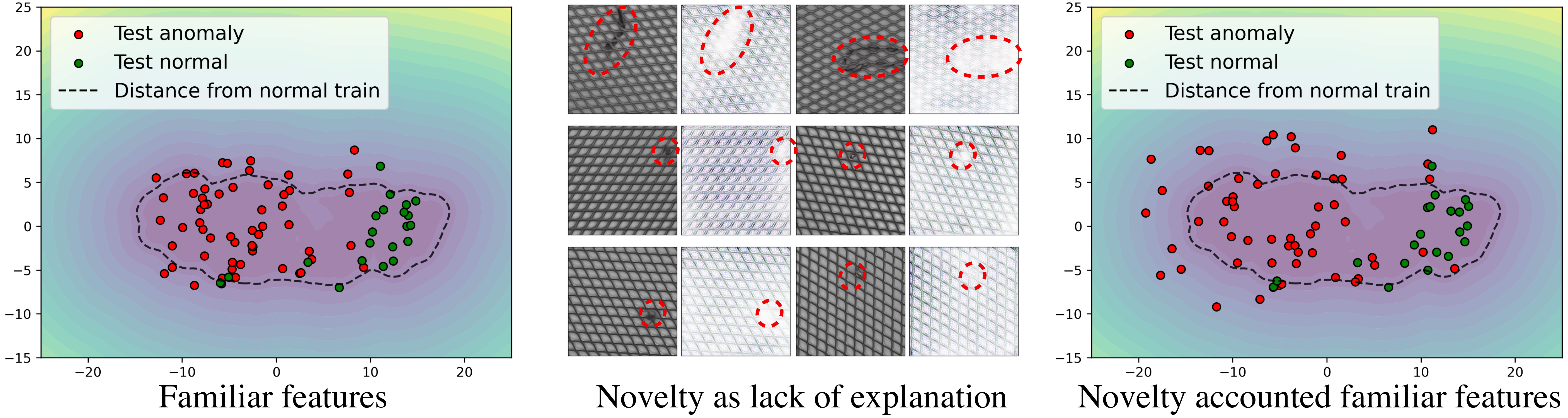}
  \caption{The PCA plot shows the normal test and anomaly test samples plotted on the two principal components. The first plot is the PCA using familiar features and the second is with novelty score added to the features. The contour shows the sum of distances to the two nearest train normal samples. The samples shown are the ones that give maximum deviation from train normal on adding novelty score.}
  \label{fig:fig10}
\end{figure*}

Figure~\ref{fig:fig10} shows further analysis of the reduction of false negatives in a challenging class of the MVTec AD dataset. It shows the features of normal test samples and test anomalies in their principal component space. The second plot is the features added with the sign-corrected novelty score. The color of the contour at each point shows the sum of the distances to the two nearest normal train samples. The samples and explanation are the ones that showed the highest difference by incorporation of novelty into the computation. The evidence of novel features is captured by Equation~\ref{Novelty}. The PCA plots show how accounting for novelty moves the anomaly samples further away from the normal train compared to the normal test samples. The figure shows how accounting for novelty as lack of explanation helps reduce false negatives.

\begin{figure}[h]
  \centering
    \includegraphics[width=0.9\linewidth]{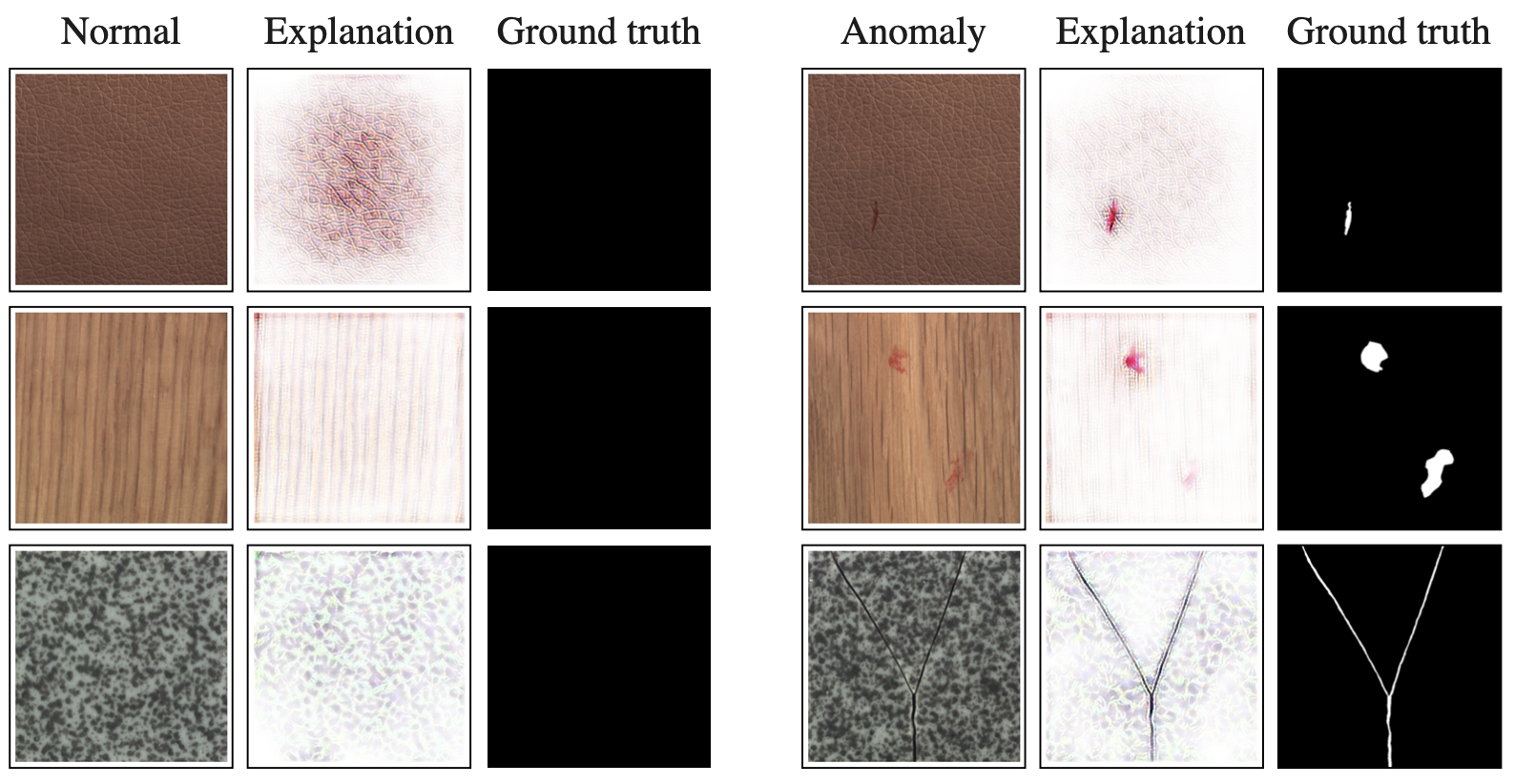}
  \caption{Test samples from MVTecAD dataset, and the explanation for being classified as normal using a $\operatorname{B-cos}$ model.}

  \label{fig:result1}
\end{figure}

\subsection{Benchmarking ENS for sensory AD}
In this experiment, we evaluate the efficacy of using the explanation of the input for sensory AD. First, we use the MVtec dataset~\cite{mvtech}, which has anomaly at the pixel level. 

\vspace{-0.2cm}

\begin{wrapfigure}{r}{0.5\textwidth}
  \centering
    \includegraphics[width=0.9\linewidth]{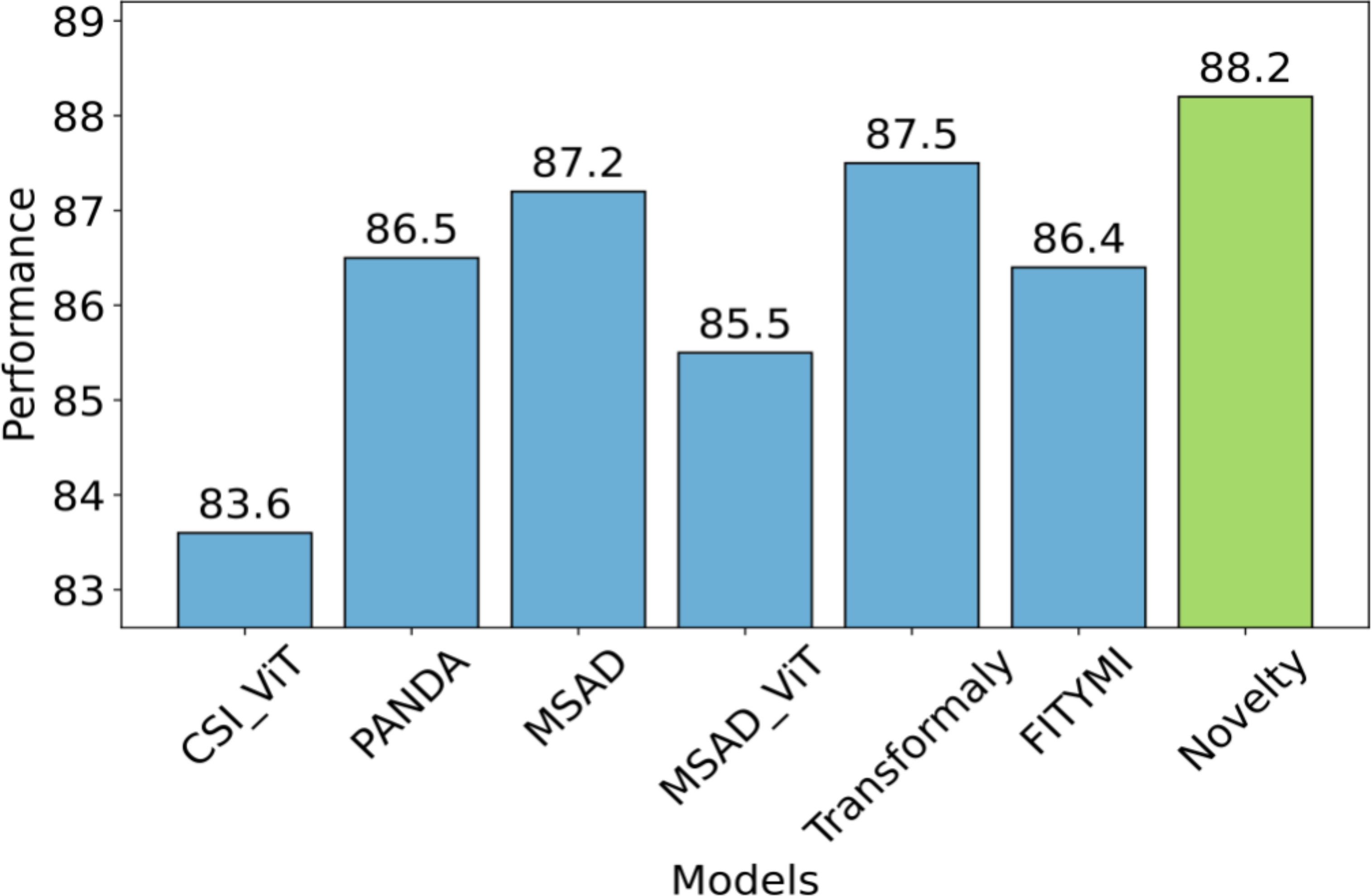}
    \caption{Performance of AD models tuned for semantic AD on MVTec AD dataset compared against novelty based scoring (ENS). }
    \vspace{-0.5cm}
    \label{fig:ENS}
\end{wrapfigure}
The prior art that considers only feature familiarity without dense matching has reported relatively lower performance in this task compared to its performance on other benchmarks. The MVtec dataset has pixel-precise annotations of all anomalies to compare the explanations.

For every normal class in the dataset, we fine-tune a $\operatorname{B-cos}$ ViT backbone pre-trained on the ImageNet-1K dataset, with a two-class classification head (one for normal and the other for outlier class). Unless mentioned otherwise, the value of B used in $\operatorname{B-cos}$ network across all experiments is $1.5$. We use data samples from the normal approximation of the normal class as the outlier. Images of the class follow a complex distribution, and a normal distribution cannot capture this complexity. The classifier is trained to discriminate this error in approximation. More details of this choice are discussed in the section~\ref{OCC}. We use standard training procedures as in ~\cite{mirzaei2023fake} without any augmentation for fine-tuning the classifier.

For a test input $x_{\mathrm{test}}$, explanation for the classification is computed as described in Section~\ref{ENS}:  ($\theta_{1\rightarrow L}(x_{\mathrm{test}})$). Figure~\ref{fig:result1} shows the explanation in the input generated by the $\operatorname{B-cos}$ model for the anomaly branch of the classification head. This is the explanation computed for the decision that the input sample is an anomaly. This shows how the method can not only detect but also explain the anomaly. We compute the ENS score on MVTec and report the same as the anomaly score for comparison with other familiarity based methods (Figure ~\ref{fig:ENS}). This improved performance comes without using feature representations of normal samples. That is, the performance improvement comes with a reduced memory (the memory to store train normal features) and computation cost (of computing the K nearest neighbour).

\begin{table*}[t]
    \centering
    \begin{tabular}{lcccccccccc}
        \toprule
        \textbf{Class} & \textbf{plane} & \textbf{Auto} & \textbf{Bird} & \textbf{Cat} & \textbf{Deer} & \textbf{Dog} & \textbf{Frog} & \textbf{Horse} & \textbf{Ship} & \textbf{Truck} \\
        \midrule
        \textbf{FFS Acc} & 0.95 & 0.84 & 0.87 & 0.90 & 0.89 & 0.78 & 0.92 & 0.96 & 0.92 & 0.83 \\
        \textbf{FFS} & \includegraphics[width=0.05\textwidth]{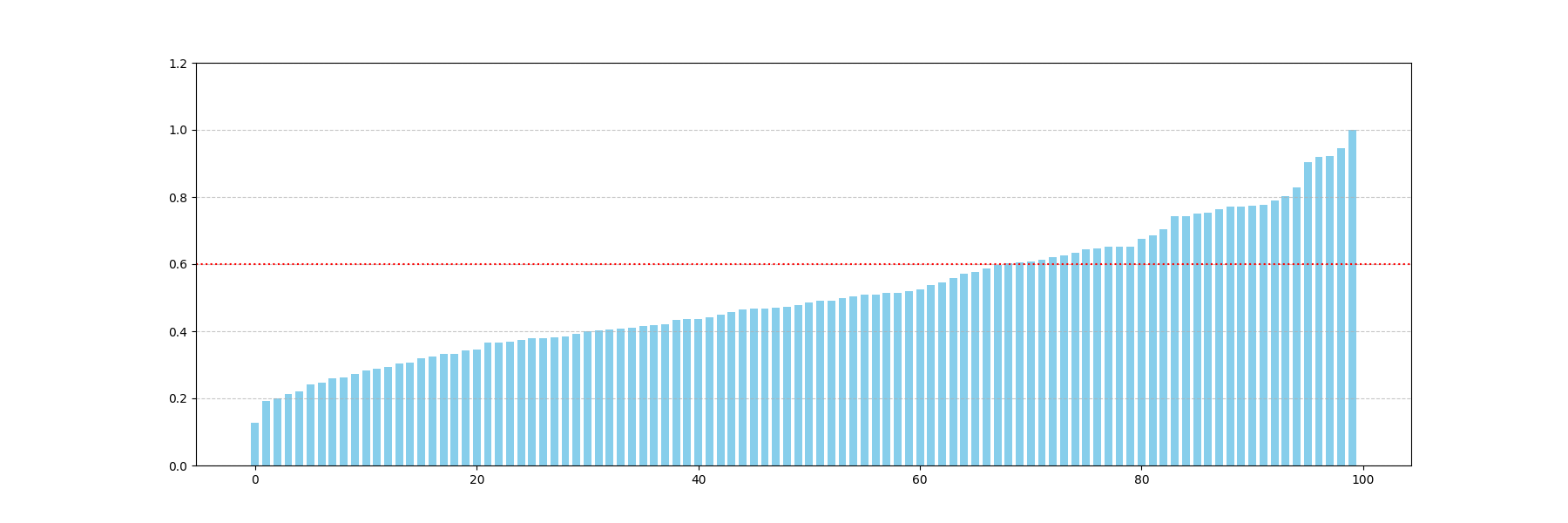} & \includegraphics[width=0.05\textwidth]{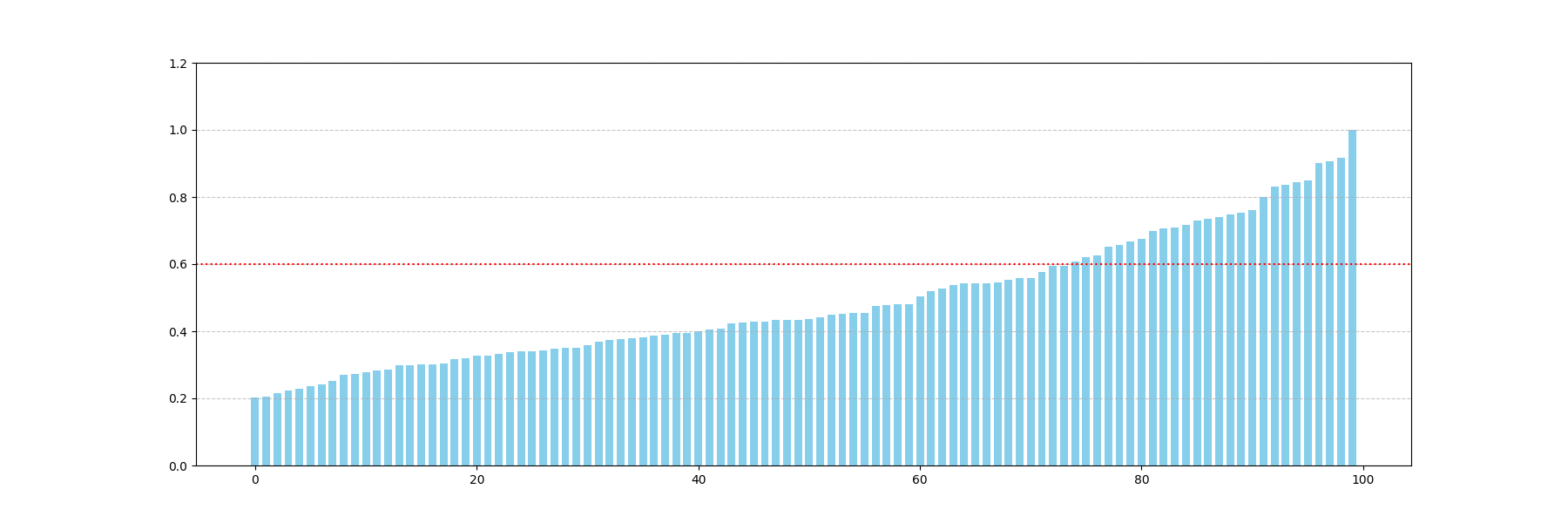} & \includegraphics[width=0.05\textwidth]{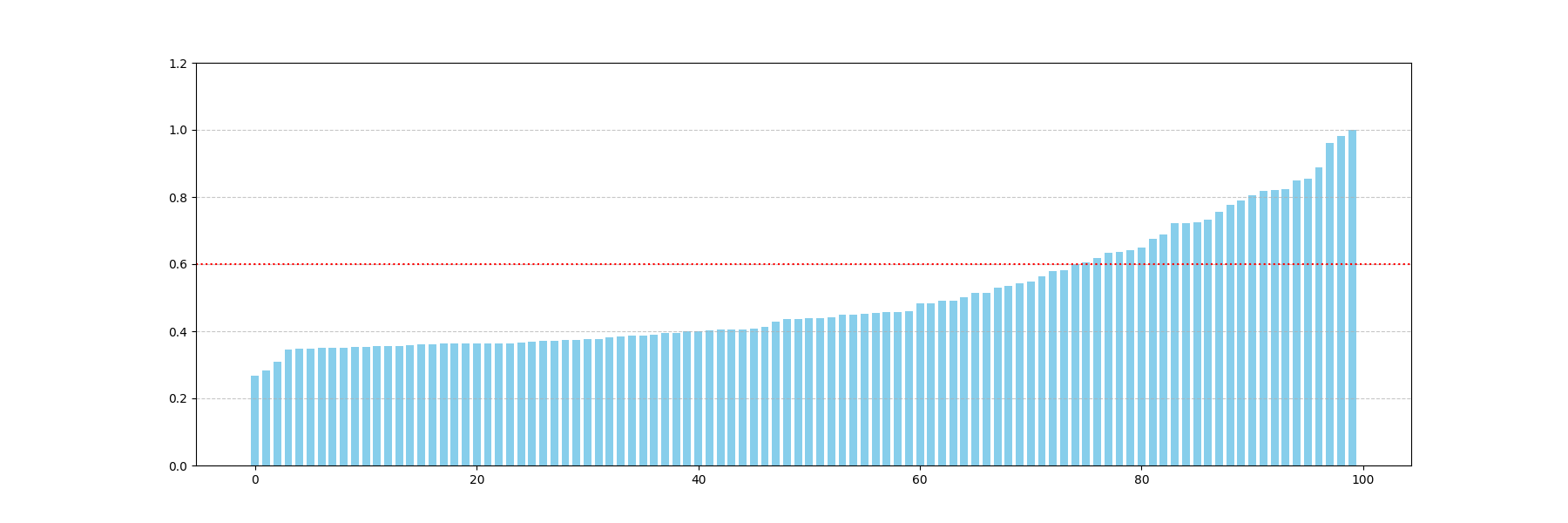} & \includegraphics[width=0.05\textwidth]{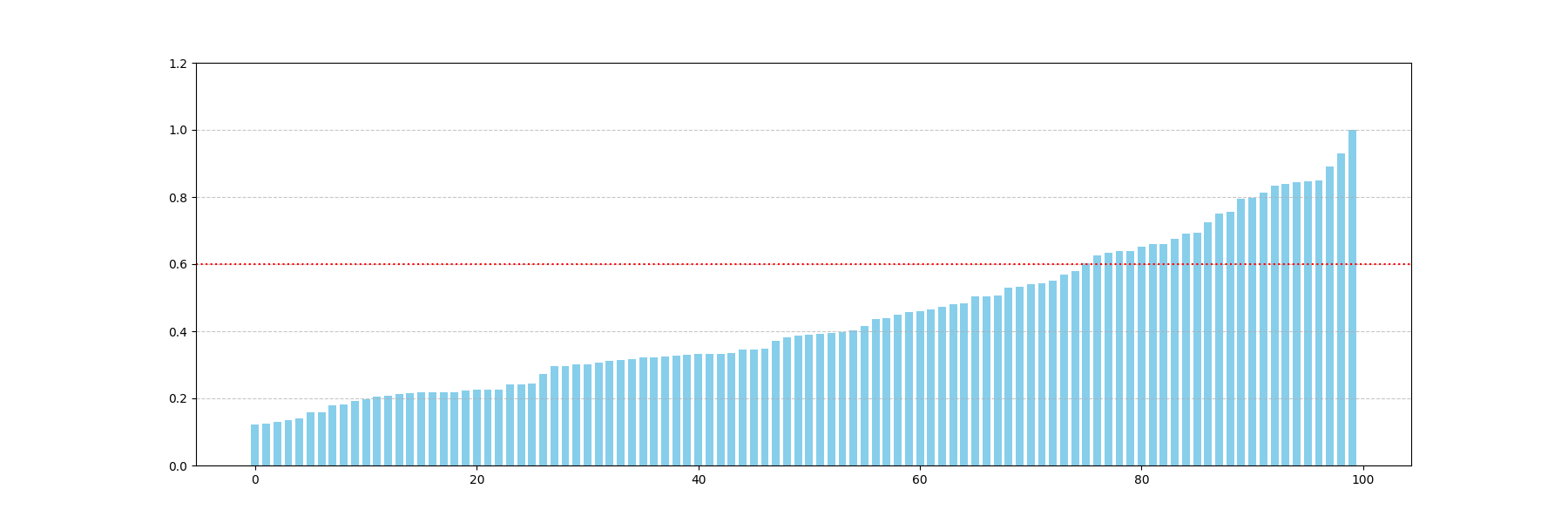} & \includegraphics[width=0.05\textwidth]{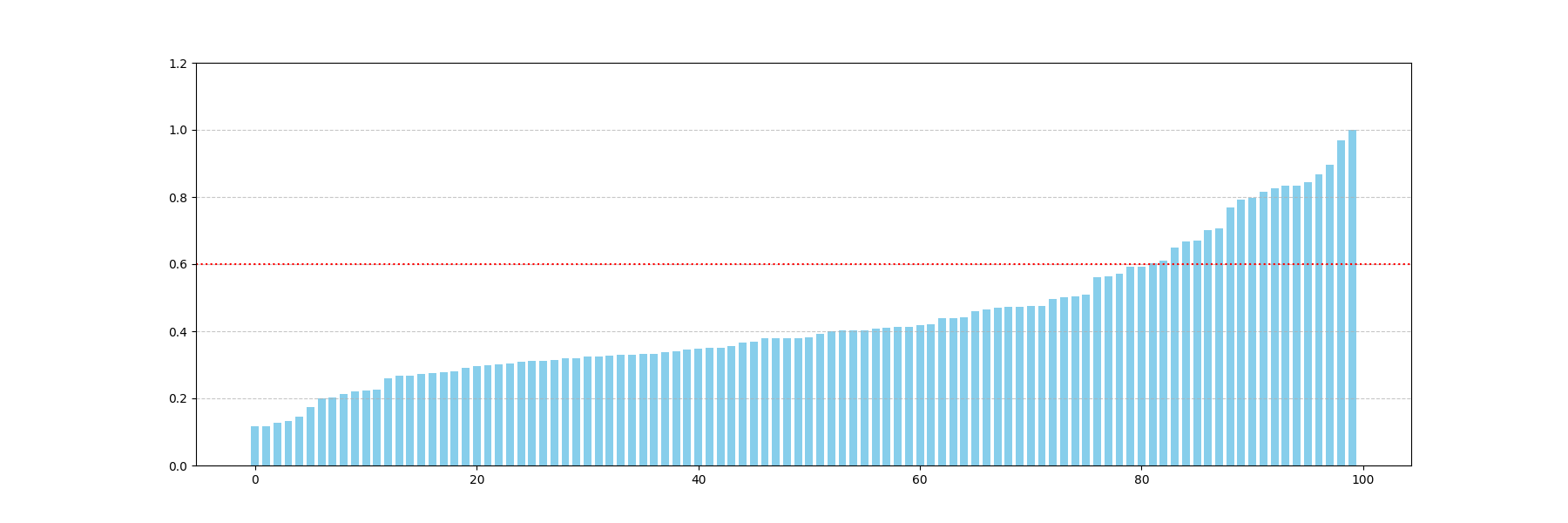} & \includegraphics[width=0.05\textwidth]{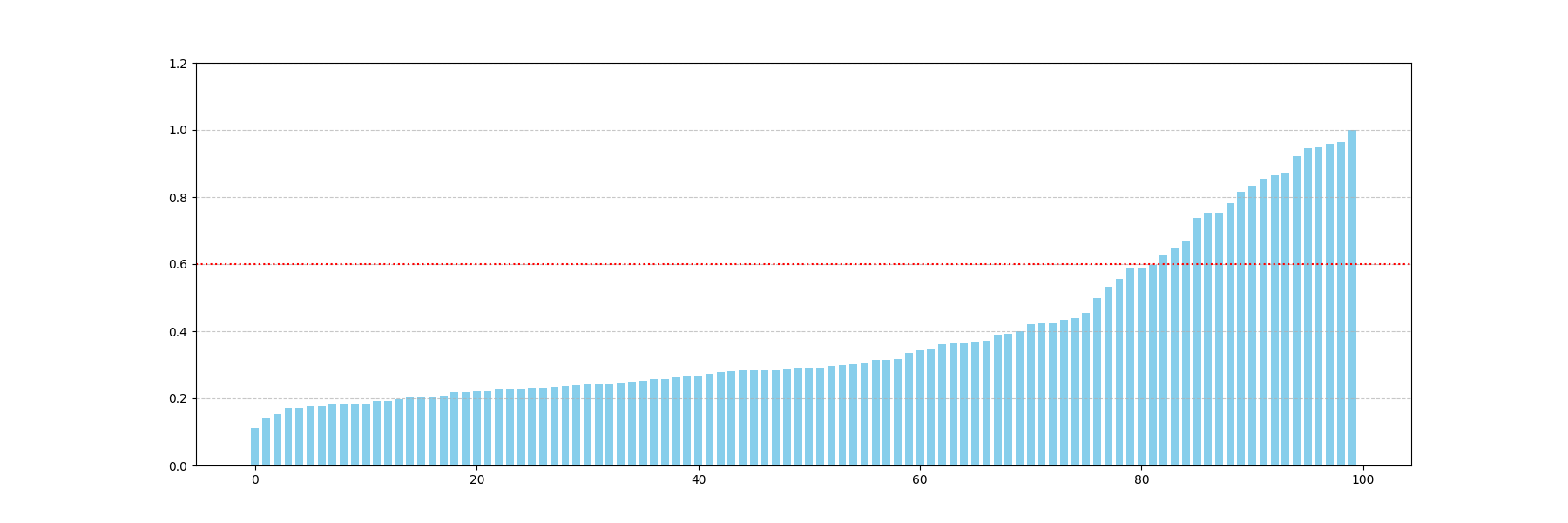} & \includegraphics[width=0.05\textwidth]{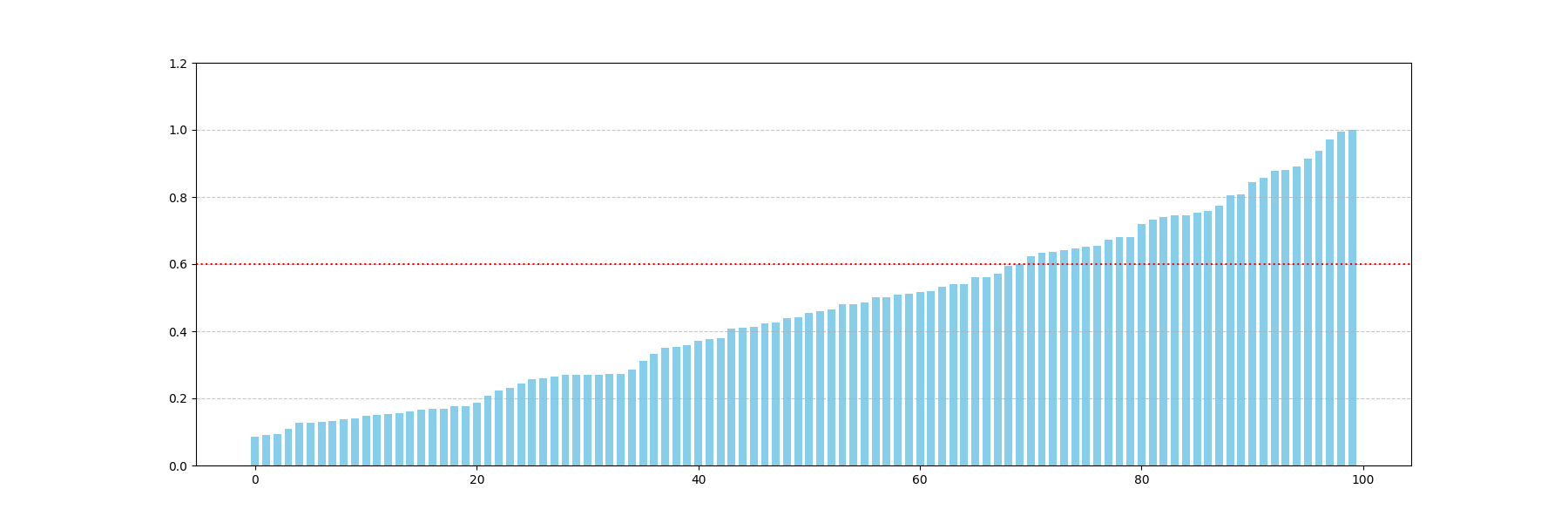} & \includegraphics[width=0.05\textwidth]{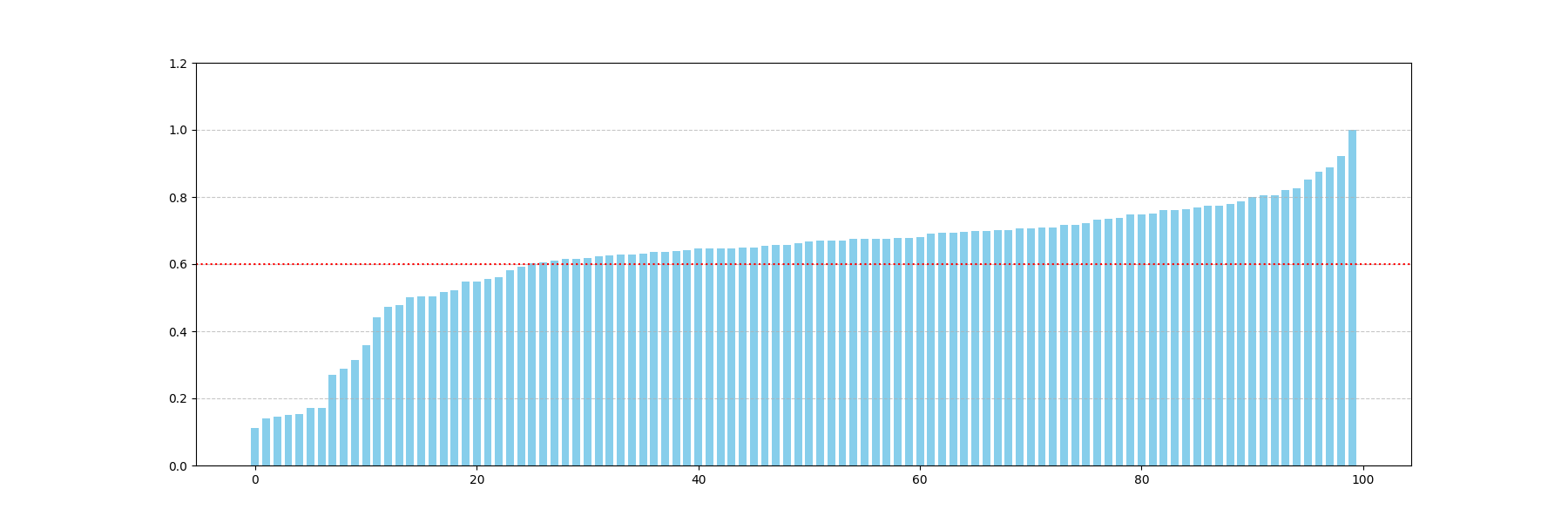} & \includegraphics[width=0.05\textwidth]{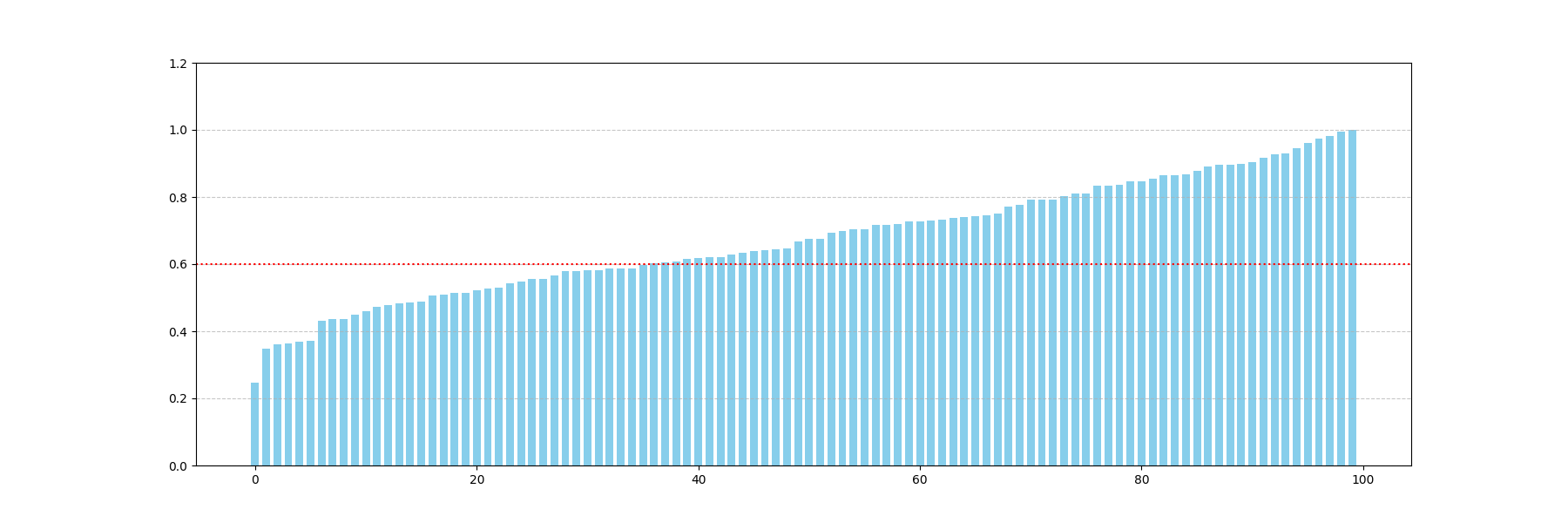} & \includegraphics[width=0.05\textwidth]{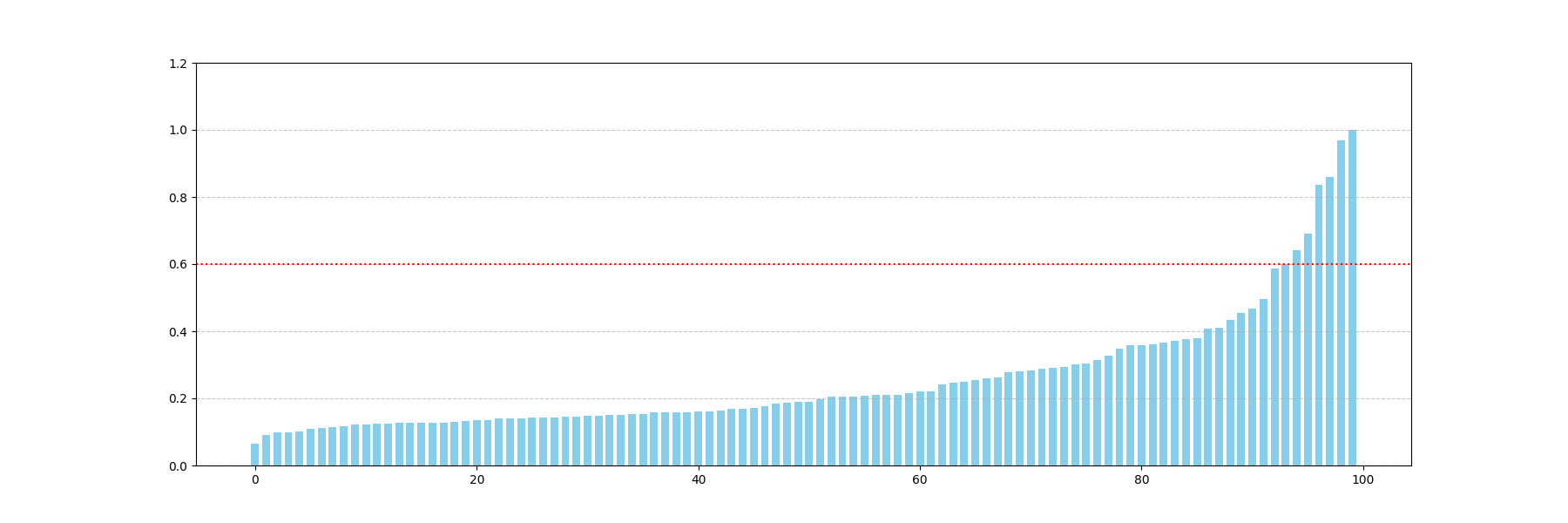} \\
        
        \textbf{ENS}  & \includegraphics[width=0.05\textwidth]{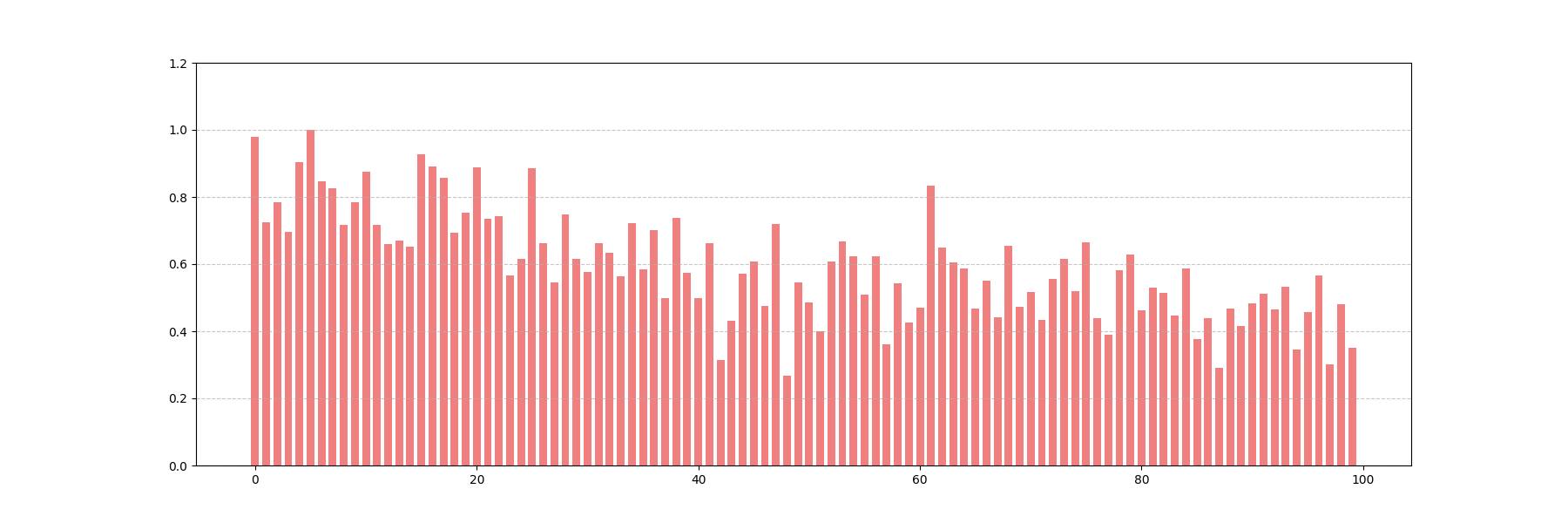} & \includegraphics[width=0.05\textwidth]{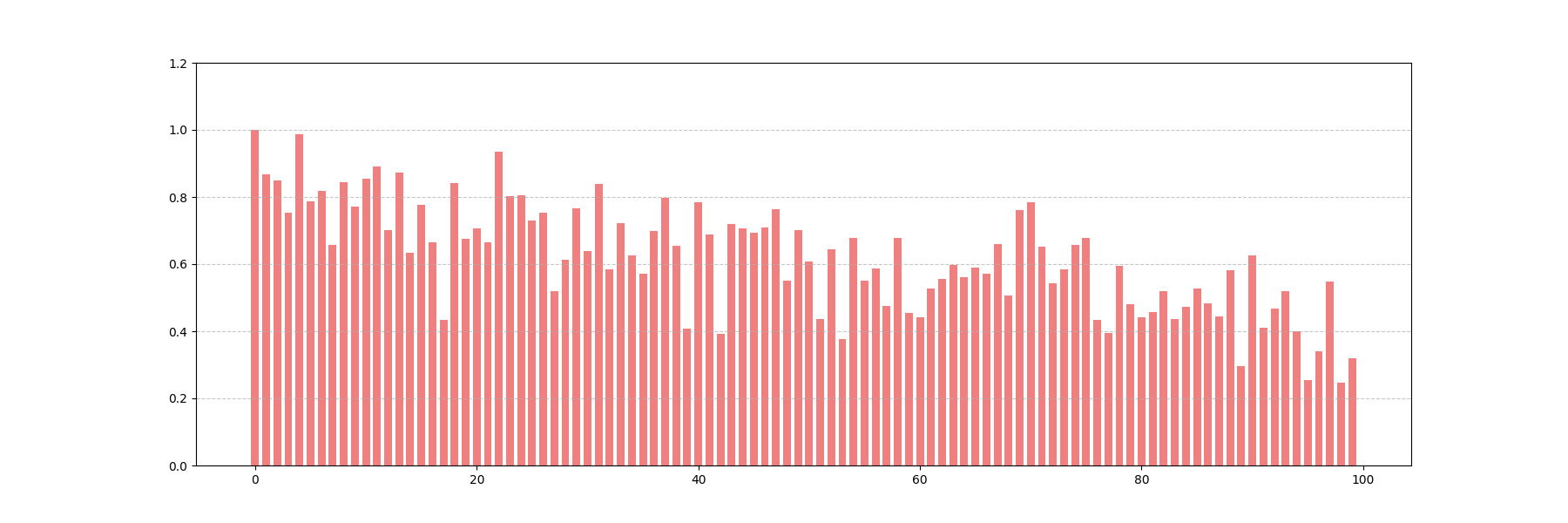} & \includegraphics[width=0.05\textwidth]{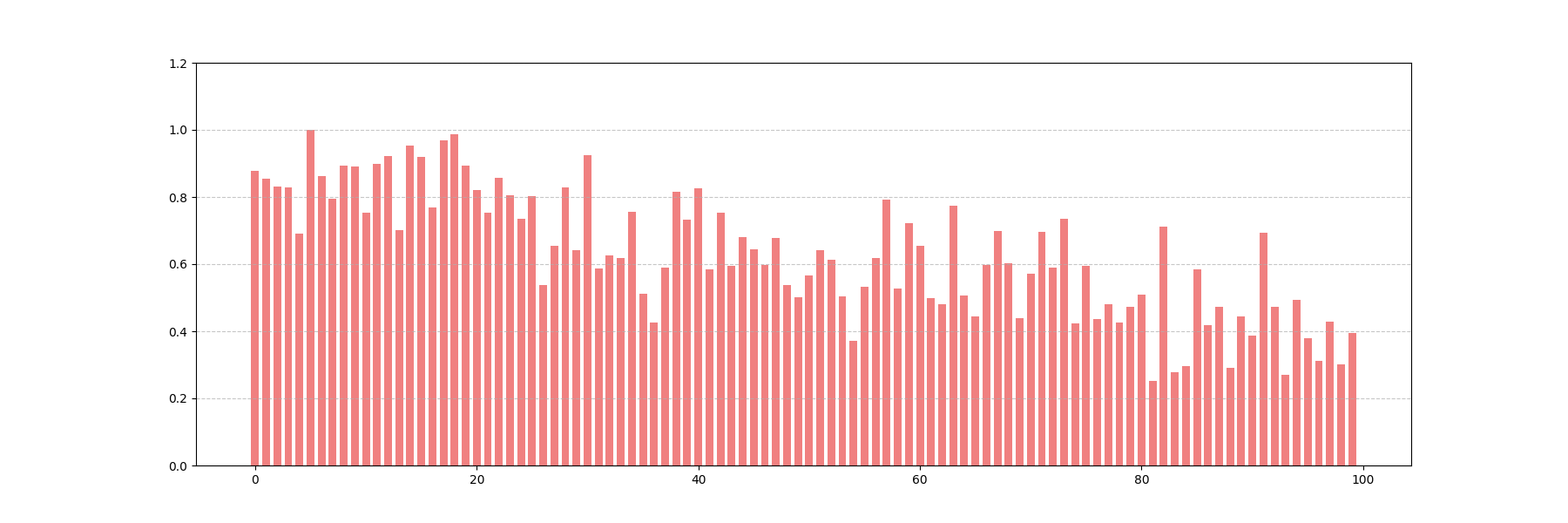} & \includegraphics[width=0.05\textwidth]{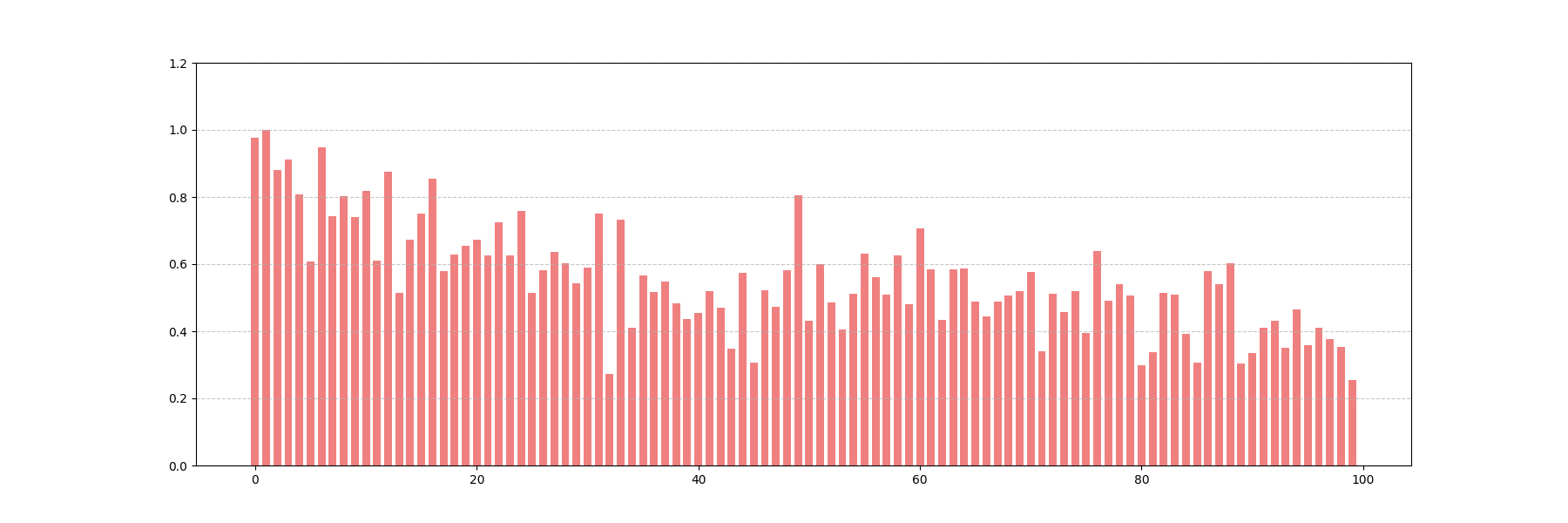} & \includegraphics[width=0.05\textwidth]{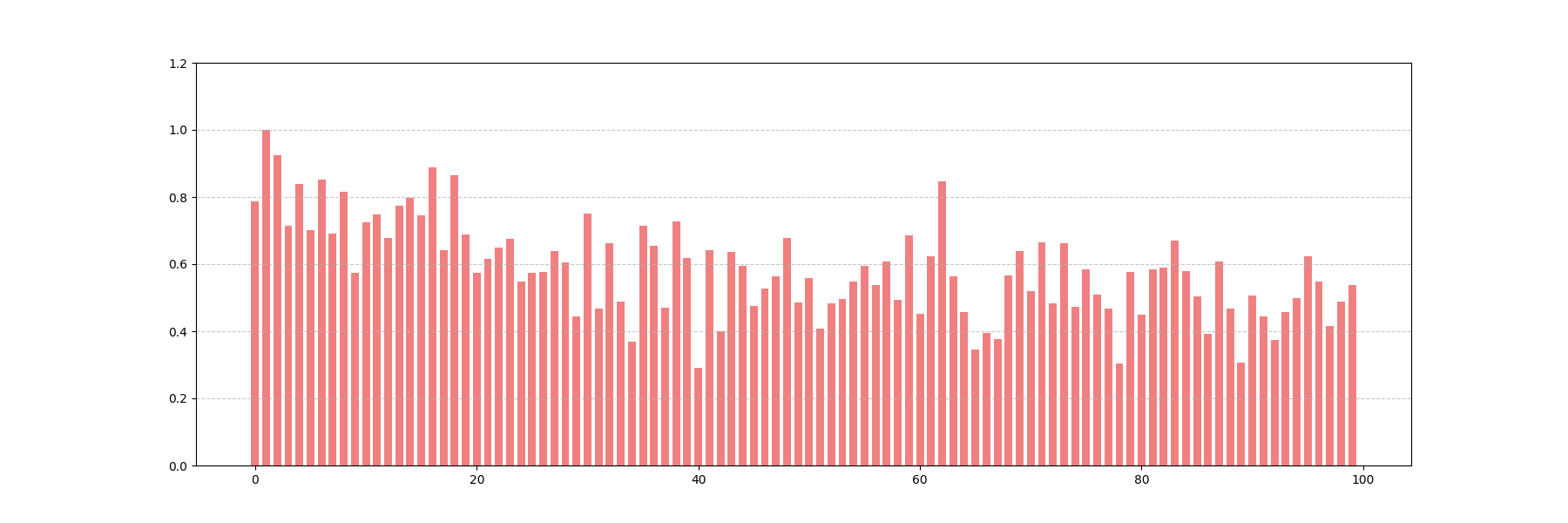} & \includegraphics[width=0.05\textwidth]{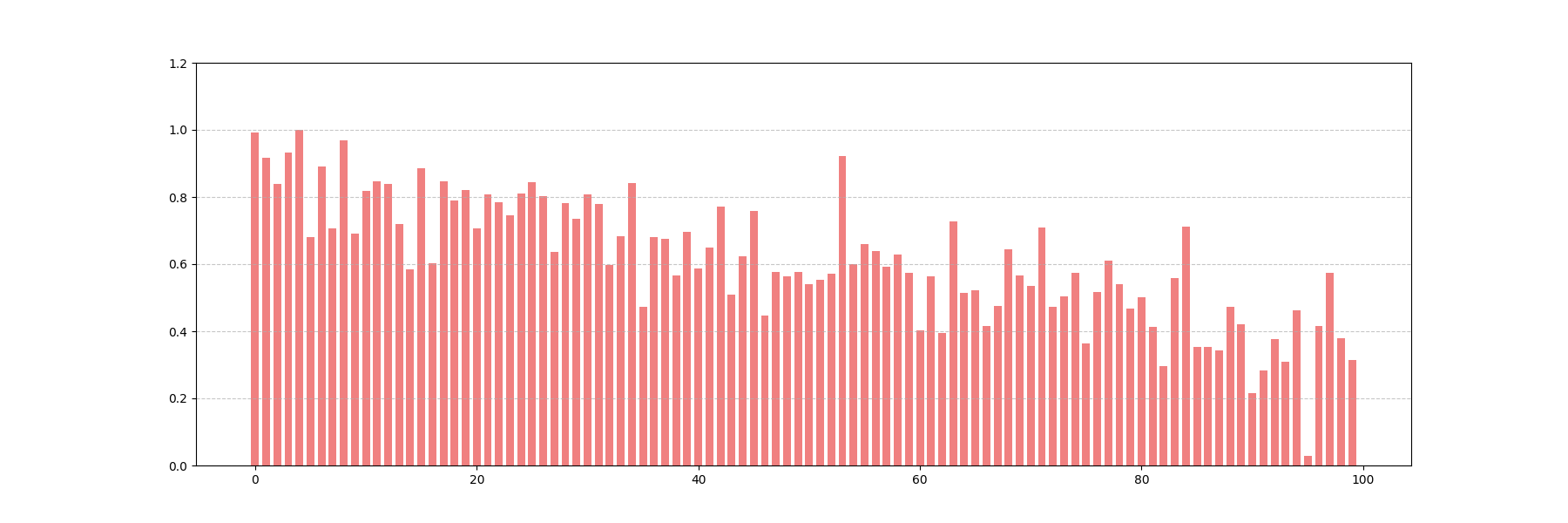} & \includegraphics[width=0.05\textwidth]{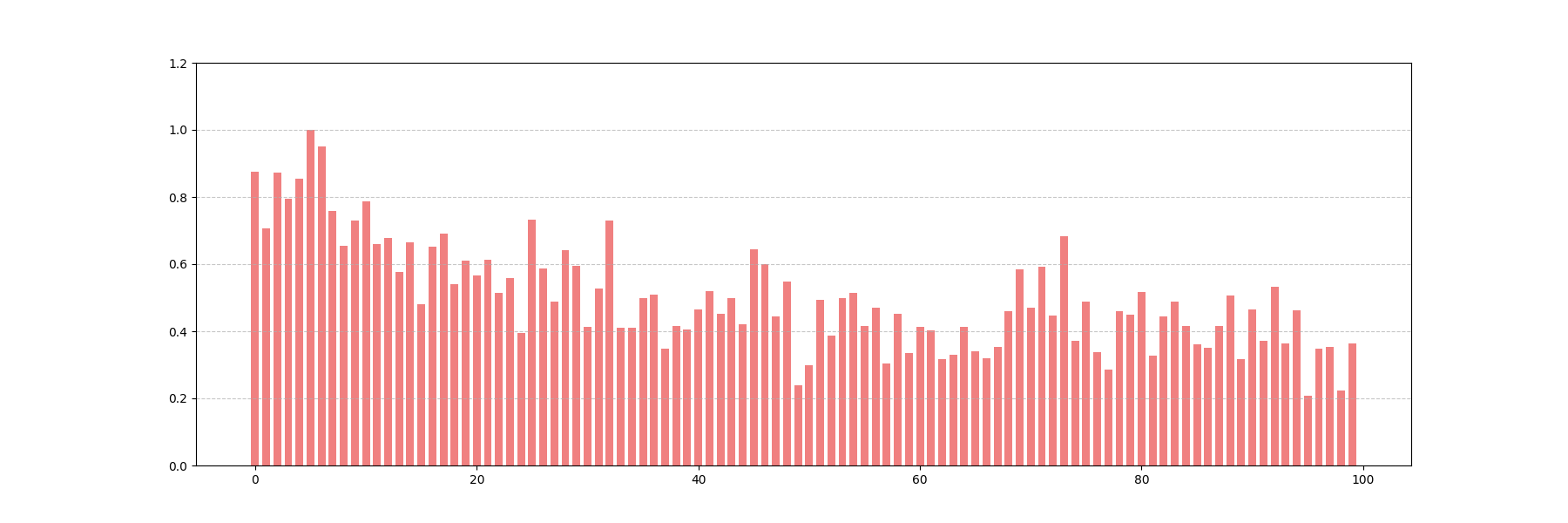} & \includegraphics[width=0.05\textwidth]{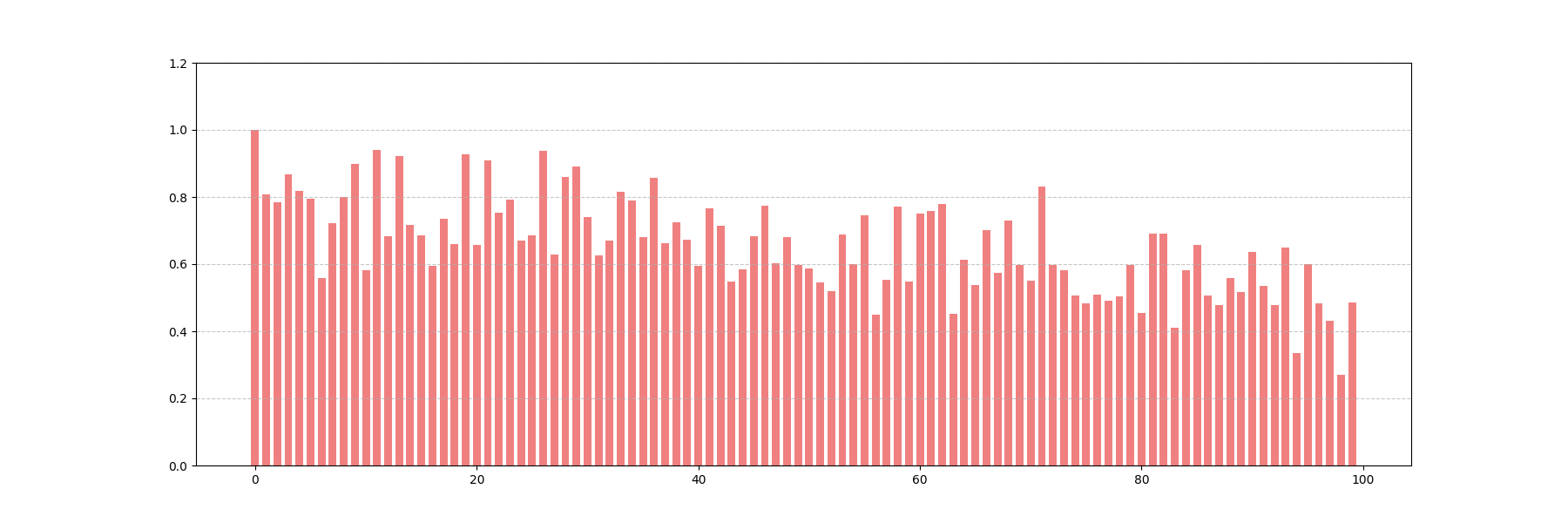} & \includegraphics[width=0.05\textwidth]{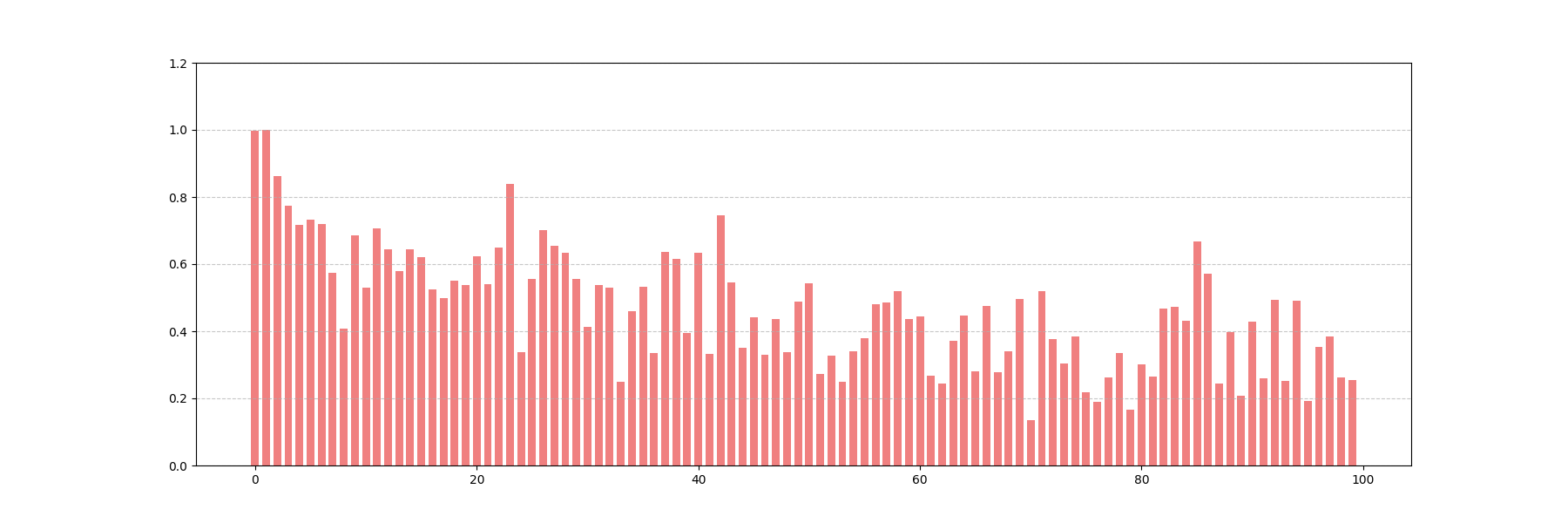} & \includegraphics[width=0.05\textwidth]{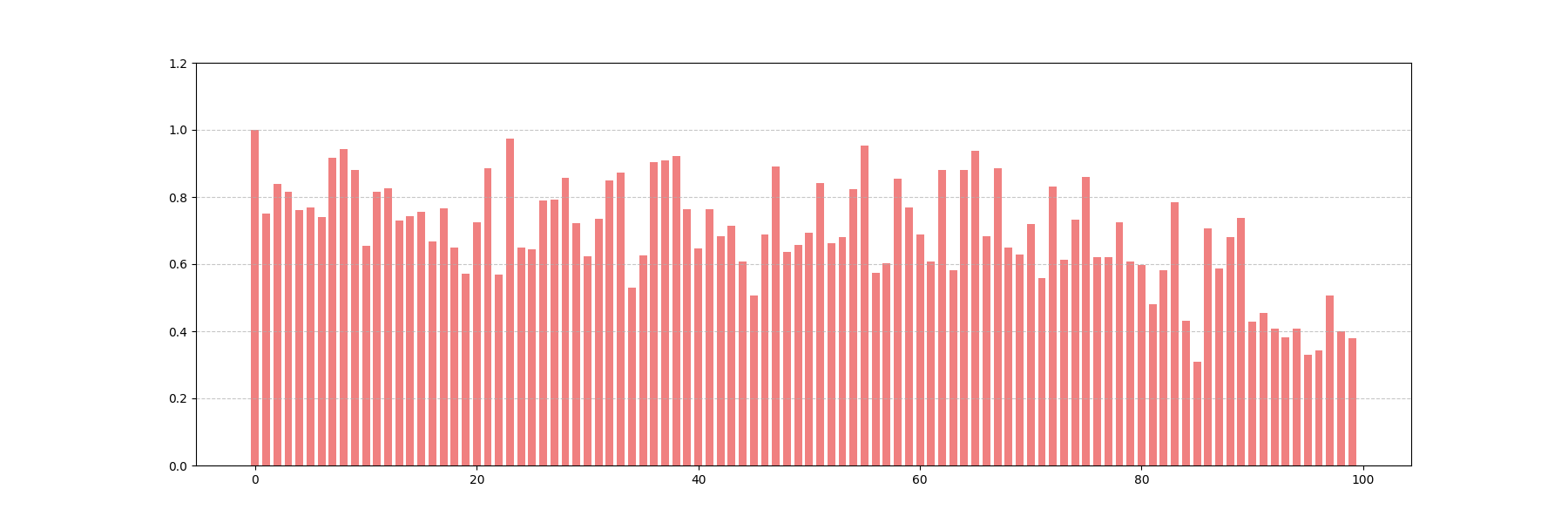} \\                \textbf{Joint model} & 0.95 & 0.85 & 0.96 & 0.90 & 0.93 & 0.91 & 0.94 & 0.97 & 0.94 & 0.83 \\

        \bottomrule
    \end{tabular}
    \caption{Ablation showing the performance of familiarity based method and the proposed joint model (FFS, and Joint Model) on CIFAR-10-100 Dataset. Table also shows the FFS and ENS scores for each of the anomaly samples in the dataset. Theentries in plot are ordered by their FFS score.}
    \label{ablation}
\end{table*}

\subsection{Reducing the role of outlier class in anomaly classification}
\label{OCC}
The impact of outlier class needs to be minimal for AD to be robust to the diverse set of anomalies encountered in the real world. Hence it is useful to have non-discriminative learning in AD. \cite{mirzaei2023fake}
uses a score-based generative model trained and prematurely early stopped on the normal samples. Appendix 1 shows evidence for the need to reduce reliance on the outlier class. 

\vspace{-0.1cm}
\begin{wrapfigure}{r}{0.5\textwidth}
  \centering
    \includegraphics[width=\linewidth]{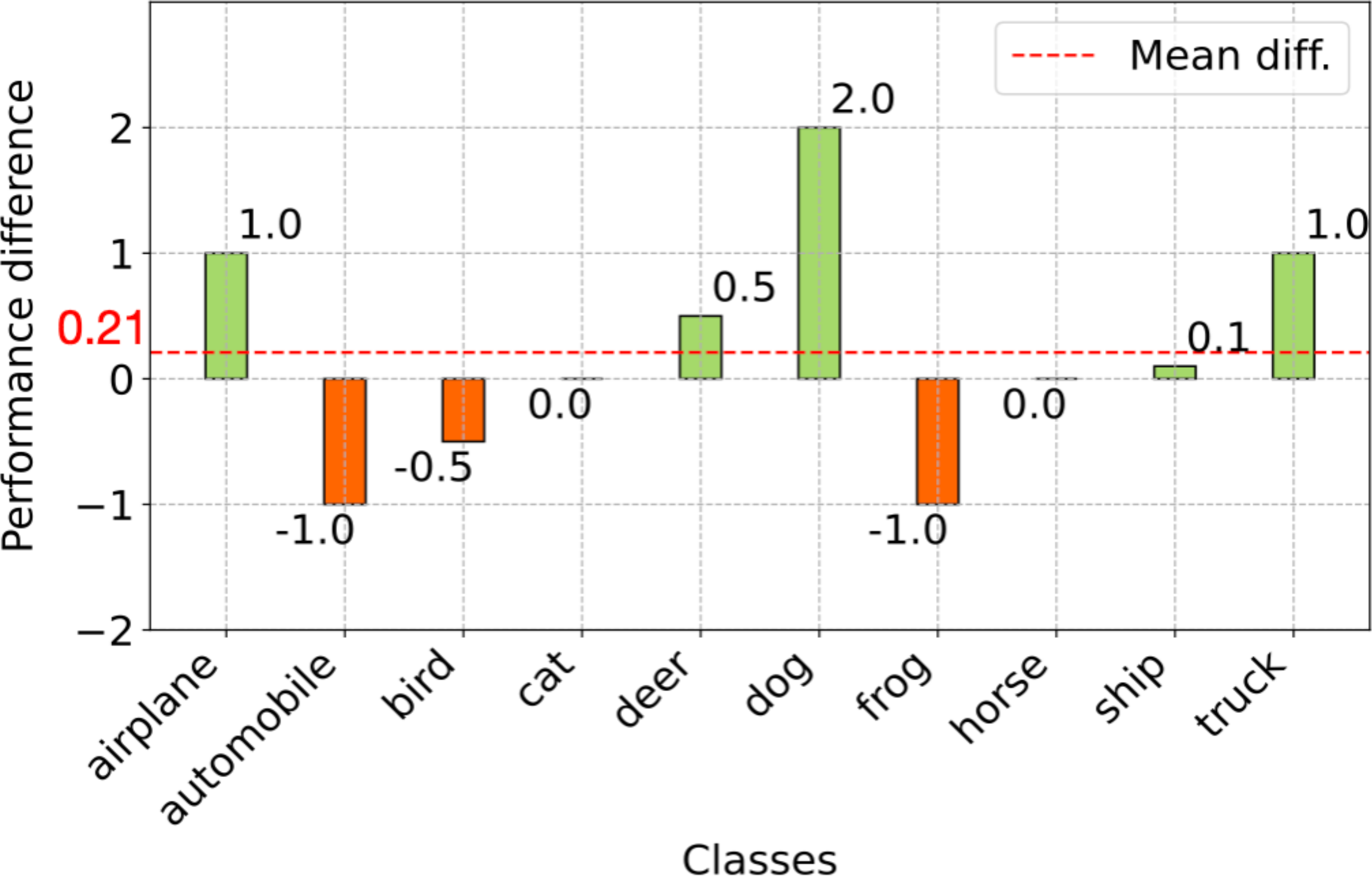}
    \vspace{-0.5cm}

    \caption{Difference in absolute AUROC using diffusion model and normal approximation for the outlier class on near semantic AD.}
    \label{fig:background}
\end{wrapfigure}

Furthermore, recent work shows adding generated images improves ID performance on standard benchmarks~\cite{diffussion}. Generated images even show promising performance in replacing real images for training image classification tasks~\cite{sariyildiz2023fake}. Stopping the generator training at the right point is vital to ensure the outlier samples are not part of the distribution.

We use a simple normal approximation of the normal class. That is, we compute the mean and covariance of the normal dataset and sample from that distribution to generate an anomaly. Using normal approximation actually gives a mean improvement of about 0.2\%, making it the preferred alternative. Figure~\ref{fig:background} shows the effect of the two different outliers in AD.

\subsection{Ablation showing the effect of FFS and ENS in AD}

In previous sections, we benchmark the effectiveness of the joint model for AD. In this section, we present an ablation study showing the effectiveness of ENS and FFS in computing the anomaly score. For this, we use the classes of the challenging semantic AD dataset Cifar10-100. For semantic AD, the novelty score is computed at the later layer of the neural network to capture higher-order semantic novel features. The Table~\ref{ablation} shows that the Joint model matches or outperforms the FFS-based methods.

The FFS score and ENS score for all the test anomalies are shown in the table as bar plots. The higher anomaly is good since all the samples shown are actually anomalies. The samples are sorted in the same order (by the amplitude of their FFS score). It is interesting to note that ENS and FFS scores somewhat complement each other. Samples whose anomaly is caused by a lack of familiarity are captured by the FFS score. An anomaly sample gets a lower FFS score when they have a large number of familiar features bringing them near normal samples in the representation space. The novel features that are not accounted for in the computation of ENS score are captured in the joint model. Hence Joint model helps give better AD performance by accounting for both the lack of familiarity and the presence of novelty.

\section{Conclusion}

This paper highlights the need for Anomaly Detection (AD) to go beyond \textit{familiar} features and incorporate \textit{novel} features into the model. We are inspired by the  `familiarity hypothesis': AD methods that rely solely on familiar features cause consistent false negatives when anomalies are caused by truly novel features that are not well captured by the pre-trained encoding. Hence, we have proposed a method to capture truly novel features as unexplained observations and show that accounting for them reduces false negatives in AD. The proposed method establishes state-of-the-art results on multiple benchmarks across different anomaly types. The method also reduces the reliance on background class, allowing the use of simpler approximation in future work. We believe further research to capture novel features in test input will continue to improve anomaly detection and related tasks like Novel Class Discovery, Out-of-Class detection, and Out-of-Distribution Detection.

\bibliography{main}
\bibliographystyle{icml2024}

\clearpage
\section{Appendix}

\subsection{Reliance on background class}

\begin{figure}[h]
    \centering
    \includegraphics[width=\linewidth]{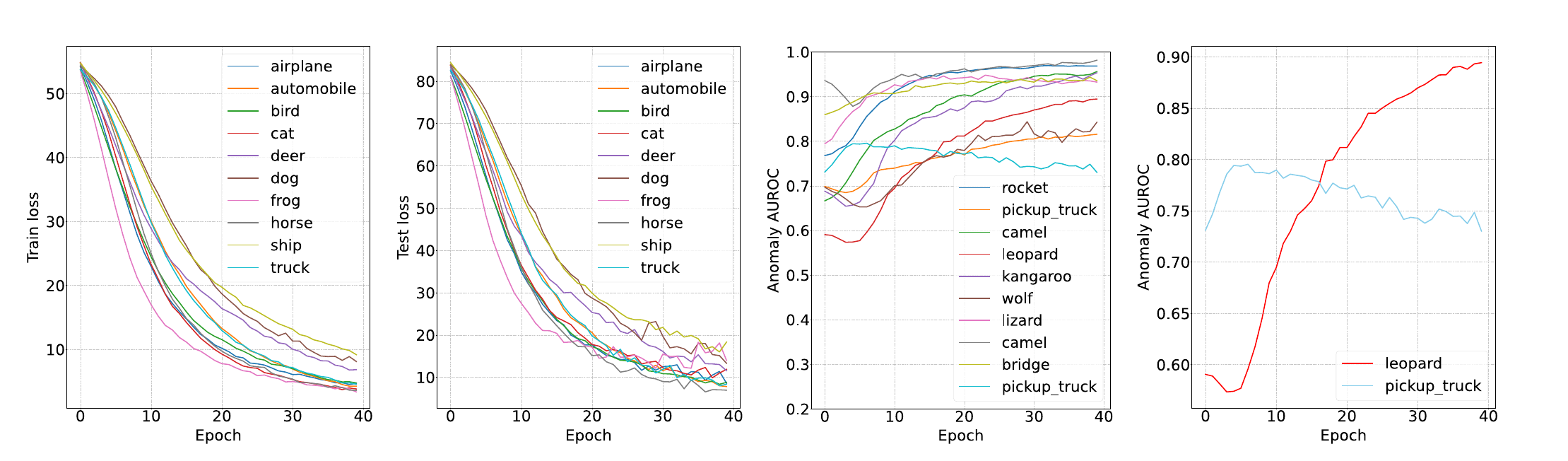}
    \caption{Training profiles of models on the C10-100 dataset. From the left, the first plot shows the training profile loss. The second plot shows the test loss. The third plot shows the training accuracy, and the fourth plot shows the test accuracy of two of the classes showing different behaviours.}
    \label{fig:app}
\end{figure}

Figure~\ref{fig:app} shows the training profile of anomaly detection on the challenging formulation of Cifar-10. For each class of Cifar-10 the most challenging class from Cifar-100 is selected as test anomaly. The training is done following the method in ~\cite{mirzaei2023fake}, such that for each normal class, we train a ViT backbone to discriminate the normal sample from samples generated with a diffusion model trained to approximate the normal distribution. 

The first two plots show the train and test loss for each of the ten classes in Cifar-10. The train and test loss shows the successful reduction of empirical risk in the classification task for which the encoder is getting trained. The third plot shows the AD performance on the held-out test anomaly. We can see that for some anomaly classes like `lizard'. the AD performance improves with the reduction of train and test losses.

Considering the anomaly classes shown separately in the fourth plot, we can clearly see how, for some anomalies, like class `pick-up truck', the AD performance reduces with improved training. Despite the reduced loss at the classification task, the AD performance reduces. This shows the poor alignment between the representation learning task and the AD task. Hence, designing background classes for any normal sample without making assumptions on the nature of anomaly is a challenging task. Reducing the reliance on background classes for feature learning is a desirable property for AD methods.

\subsection{Cifar-10 vs 100 near ND semantic anomaly Evaluation}

In earlier AD benchmark, where the classes from different dataset are considered anomalies, a Cifar-10 vs Cifar100 benchmark follow that: category within CIFAR-10 is identified as the standard or normal class, while all other categories are classified as far anomalies. ~\cite{mirzaei2023fake} presents a near AD evaluation, where the class from CIFAR-100 closest to normal class in CIFAR-10is used for evaluating detection process. It's the class associated with the lowest test AUROC in anomaly detection. 

As mentioned in the 3rd para of section 5.2, we use this framework to evaluate the near ND benchmark. The closest classes of CIFAR-100 is the anomaly corresponding to each class of CIFAR-10. To evaluate a model on this benchmark, first, the model is trained on each of the classes of Cifar-10 as a normal class. The model trained on a class, say class 'cat' from CIFAR-10, uses all samples from the train split of class 'cat' as normal for training. The model is then evaluated on all classes of CIFAR-100 one after the other, each as an anomaly with the test samples of class 'cat' as test normal samples. Then, the most challenging class (least AUROC) from CIFAR-100 is selected as the test anomaly. This is repeated for all 10 classes of CIFAR-10. Correspondingly, we get a near anomaly from CIFAR-100. Below are the normal class near anomaly pairs following the format 'CIFAR-10 normal class':{CIFAR-100 anomaly class}:

C10-100 = \{'airplane': 'rocket', 'automobile': 'pickup-truck', 'bird': 'kangaroo', 'cat': 'rabbit', 'deer': 'kangaroo', 'dog': 'wolf', 'frog': 'lizard', 'horse': 'camel', 'ship': 'bridge', 'truck': 'pickup-truck'\}

The intuition here is that any other class in the Cifar-100 as an anomaly would yield a better AUROC score, making this the most challenging benchmark.

\section{Choice of K in K-nearest neighbours for FFS}

We choose K=2 for the K-nearest neighbour model for familiar feature based anomaly detection. This design choice is taken from the prior art for a fair comparison~\cite{mirzaei2023fake}. Furthermore, the method is robust to the change of the nearest neighbour to a large extent, as shown by Mirzaei~\etal.

\begin{center}
\begin{tabular}{ccc *{5}{c}}
\toprule
Experiment & Dataset & \multicolumn{5}{c}{C} \\
\midrule
 & & k=1 & k=2 & k=3 & k=4 & k=5 \\
\midrule
1 & CIFAR-10 & 99.0 & 99.1 & 98.9 & 98.9 & 98.7 \\
2 & CIFAR-10v100 & 89.8 & 90.0 & 90.0 & 90.0 & 89.7 \\
\bottomrule
\end{tabular}
\end{center}

\end{document}